\documentclass[lettersize,journal]{IEEEtran}
\usepackage{cite}
\usepackage{algorithm}
\usepackage{algpseudocode} 
\usepackage{epsfig}
\usepackage{graphicx}
\usepackage{amsfonts}
\usepackage{comment}
\usepackage{caption}
\usepackage{color}
\usepackage{amssymb}
\usepackage{amsmath,dsfont,mathrsfs,bbold}
\usepackage{amsthm,mathtools}
\usepackage{multirow}
\usepackage{etoolbox}
\usepackage{balance}
\usepackage{booktabs,adjustbox}
\usepackage{enumitem}
\usepackage[table]{xcolor}

\newcommand{\cmark}{\textcolor{green}{\ding{51}}}
\newcommand{\pmark}{\textcolor{blue}{\ding{51}}}
\newcommand{\xmark}{\textcolor{red}{\ding{55}}}%
\usepackage{pifont}

\usepackage[colorlinks = true,
    linkcolor = red,
    urlcolor  = black,
    citecolor = blue, 
    anchorcolor = magenta]{hyperref}

\usepackage{tikz}

\usepackage{caption}
\usepackage{boldline}
\usepackage{makecell}
\setcellgapes{1pt}
\usepackage{rotating}
\usepackage{lineno}

\usepackage{pgfplots}
\usepackage{subcaption}
\pgfplotsset{compat=1.17}

\makeatletter

\def\addlegendimage{\pgfplots@addlegendimage}
\newcommand*{\doubleequation}[3][]{%
    \par\vskip\abovedisplayskip\noindent
    \if\relax\detokenize{#1}\relax
       \let\@dblLabelI\@empty
       \let\@dblLabelII\@empty
    \else 
       \@dblequationAux #1,%
    \fi
    \makebox[0.5\linewidth-1.5em]{%
     \hspace{\stretch2}%
     \makebox[0pt]{$\displaystyle #2$}%
     \hspace{\stretch1}%
    }%
    \makebox[0.5\linewidth-1.5em]{%
     \hspace{\stretch1}%
     \makebox[0pt]{$\displaystyle #3$}%
     \hspace{\stretch2}%
    }%
    \makebox[3em][r]{(%
  \refstepcounter{equation}\theequation\@dblLabelI, 
  \refstepcounter{equation}\theequation\@dblLabelII)}%
  \par\vskip\belowdisplayskip
}

\makeatother
\AfterEndEnvironment{lemma}{\noindent\ignorespaces}

\newtheorem{definition}{Definition}[section]
\AfterEndEnvironment{definition}{\noindent\ignorespaces}

\AfterEndEnvironment{example}{\noindent\ignorespaces}

\ifCLASSINFOpdf
\else
\fi
\begin{document}
\title{Anomaly Resilient Temporal QoS Prediction using Hypergraph Convoluted Transformer Network}

\author{Suraj~Kumar,~\IEEEmembership{Student Member,~IEEE},
        Soumi~Chattopadhyay,~\IEEEmembership{Senior Member,~IEEE},
        Chandranath~Adak,~\IEEEmembership{Senior Member,~IEEE}

\IEEEcompsocitemizethanks{\IEEEcompsocthanksitem Suraj Kumar and Soumi Chattopadhyay are with the Dept. of CSE, Indian Institute of Technology Indore, Madhya Pradesh 452020, India.  
(email: \emph{\{phd2301101002, soumi\}@iiti.ac.in}).\\
Chandranath Adak is with the Dept. of CSE, Indian Institute of Technology Patna, Bihar 801106, India. 
(e-mail: \emph{chandranath@iitp.ac.in}).\\
Corresponding authors: S. Chattopadhyay, C. Adak.\\
S. Chattopadhyay acknowledges the partial support from IIT Indore under
YFRSG (IITI/YFRSG/2023-24/Phase-III/09).

Digital Object Identifier 10.1109/TNSM.2026.3674650
}}


\markboth{IEEE Transactions on Network and Service Management}
{IEEE Transactions on Network and Service Management}

\maketitle

\begin{abstract}
  Quality-of-Service (QoS) prediction is a critical task in the service lifecycle, enabling precise and adaptive service recommendations by anticipating performance variations over time in response to evolving network uncertainties and user preferences. 
However, contemporary QoS prediction methods frequently encounter data sparsity and cold-start issues, which hinder accurate QoS predictions and limit the ability to capture diverse user preferences.
Additionally, these methods often assume QoS data reliability, neglecting potential credibility issues such as outliers and the presence of greysheep users and services with atypical invocation patterns. Furthermore, traditional approaches fail to leverage diverse features, including domain-specific knowledge and complex higher-order patterns, essential for accurate QoS predictions.
In this paper, we introduce a real-time, trust-aware framework for temporal QoS prediction to address the aforementioned challenges, featuring an end-to-end deep architecture called the Hypergraph Convoluted Transformer Network (HCTN). 
HCTN combines a hypergraph structure with graph convolution over hyper-edges to effectively address high-sparsity issues by capturing complex, high-order correlations. 
Complementing this, the transformer network utilizes multi-head attention along with parallel 1D convolutional layers and fully connected dense blocks to capture both fine-grained and coarse-grained dynamic patterns. 
Additionally, our approach includes a sparsity-resilient solution for detecting greysheep users and services, incorporating their unique characteristics to improve prediction accuracy. Trained with a robust loss function resistant to outliers, HCTN demonstrated state-of-the-art performance on the large-scale WSDREAM-2 datasets for response time and throughput.

\end{abstract}

\begin{IEEEkeywords}
Graph convolution, Hypergraph, Temporal QoS prediction, Transformer network
\end{IEEEkeywords}

\section{Introduction}
\noindent
Over the past decade, service-oriented architectures (SOA) \cite{soa} such as Google Cloud Platform and Amazon Web Services have revolutionized software development by speeding up application creation, reducing costs, and enhancing scalability. This evolution has led to the widespread use of web services for low-cost communication and has driven the Everything-as-a-Service paradigm \cite{xaas}, prompting businesses to adopt SOA across sectors such as food, healthcare, and transportation. However, with numerous similar services available, users often struggle to choose the best option, often relying on past experiences or external opinions, a behavior known as the bandwagon effect \cite{Bandwagon_Effect}. The incomplete information about available options exacerbates this issue. Researchers emphasize the importance of non-functional parameters, collectively referred to as \textit{Quality of Service} (QoS) \cite{UPCC, IPCC, WSPRED}, for distinguishing between services. QoS can vary due to factors like network conditions, invocation time, and contextual elements. Manually assessing the QoS of millions of services is impractical and costly, underscoring the need for QoS prediction tools to assist in service recommendation.

Researchers have addressed service selection via automatic QoS prediction, with Collaborative Filtering (CF) being a prominent method that identifies correlations between users and services \cite{Survey2-TSC, Survey1-TSC_Zheng}. CF is divided into Memory-based and Model-based approaches. Memory-based CF estimates QoS by calculating similarities but suffers from high computational cost and data sparsity \cite{UPCC, IPCC, WSRec}. Model-based CF methods like Matrix Factorization (MF) \cite{NMF, NIMF}, Factorization Machines (FM) \cite{EFM}, and deep learning models \cite{DeepTSQP, RNCF, OFFDQ} handle sparsity by learning latent features, with FM capturing second-order interactions and deep learning models addressing higher-order features. Hybrid models combine both to improve accuracy \cite{OFFDQ}. However, many assume static QoS parameters \cite{NIMF, hsanet}, though QoS can fluctuate over time. While some methods model temporal QoS, they struggle with data anomalies, 
leading to unreliable predictions \cite{WSPRED, CTF, tpmcf, DeepTSQP}.


We identify three key factors causing accuracy degradation in temporal QoS prediction: 
(a) \textit{Data Deficiency}: Sparse QoS experiences, especially in cold-start scenarios, arise from limited user-service interactions \cite{BGCL}. 
(b) \textit{Data Credibility}: Dynamic environments introduce outliers, while the greysheep problem complicates predictions for atypical instances with unique traits. 
(c) \textit{Data Learning}: Traditional methods using similarity features \cite{UPCC, IPCC} or linear matrix decomposition \cite{NMF, NIMF} fail to capture complex user-service relationships over time. While some approaches leverage GCNs \cite{BGCL, QoSGNN} enhance feature learning, they struggle with temporal dynamics, and models like ARIMA \cite{TASR}, RNNs \cite{DeepTSQP, dgncl}, and Transformers \cite{tpmcf} enhance temporal feature learning, they often overlook local spatial context and struggle with data anomalies, impacting accuracy.
A recent study, TPMCF \cite{tpmcf}, employs distinct architectures to separately capture spatial and temporal features but faces challenges in managing certain data anomalies, which impacts its prediction accuracy.

Considering the abovementioned extremity, we propose an anomaly-resilient temporal QoS prediction framework called Hypergraph Convoluted Transformer Network (HCTN), an end-to-end architecture with five key components:
(a) The first module applies non-negative matrix decomposition to extract latent user and service features, addressing data sparsity and cold-start issues.
(b) The second module uses hypergraph collaborative filtering to capture high-order user-service features through hypergraph convolution over hyperedges, further addressing sparsity.
(c) To tackle the greysheep problem, the third module emphasizes local characteristics of greysheep users and services, enhancing the effectiveness of collaborative filtering.
(d) The next module refines temporal granularity by feeding updated embeddings into an enhanced transformer network that combines multi-head attention \cite{TRANSFORMER} with parallel 1D-convolutional layers and fully connected networks. This enables the model to capture fine- and coarse-grained temporal patterns, providing a rich representation of both short-term variations and long-term trends.
(e) The final module predicts QoS by integrating collaborative, spatial, and temporal features, with the entire model trained end-to-end using an outlier-resilient loss function to handle anomalies.
This framework effectively addresses data sparsity, cold-start, greysheep, and temporal dynamics challenges, providing accurate QoS predictions. 
In summary, this paper has following \textbf{contributions}:

{\emph{(i) Novelty in model architecture}}: We introduce the Hypergraph Convoluted Transformer Network (HCTN) for temporal QoS prediction. HCTN combines hypergraph learning for higher-order collaborative filtering with dual-channel Transformer networks, capturing dynamic features from fine-to-coarse-grained levels and using multi-head attention to manage various contexts.

{\emph{(ii) Handling data deficiency}}: We address data sparsity using higher-order collaborative filtering with diverse hyperedges in second-order user-service graphs. Additionally, non-negative matrix decomposition uncovers hidden user and service preferences, improving cold-start performance and enhancing collaborative filtering features.

{\emph{(iii) Tackling data credibility}}: 
To address data credibility, we apply two sparsity-resilient unsupervised algorithms to detect outliers and greysheep. Outliers are handled with a logarithmic penalty on training errors, while greysheep are managed by prioritizing their unique patterns over global collaborative filtering features, improving QoS prediction.

{\emph{(iv) Rich data representation}}: In our framework, domain knowledge features, such as latent user preferences and service characteristics, are combined with higher-order complex spatio-temporal features to effectively capture the intricate triadic relationships among users, services, and time, thereby enhancing overall data representation.

{\emph{(v) Extensive experiments}}: We conducted comprehensive experiments using the QoS benchmark dataset, WSDREAM-2 \cite{WSDREAM}, for two different QoS parameters, response time and throughput. These experiments highlight the superior performance of HCTN compared to contemporary methods. Additionally, our ablation study validates the effectiveness of each module within the proposed framework.

Section \ref{sec:problem_formulation} formulates the temporal QoS prediction problem. Section \ref{sec:proposed_method} details our framework. Section \ref{sec:experiments} presents the experiments and comparative studies. Section \ref{sec:lit_review} reviews existing works. Finally, Section \ref{sec:conclusion} concludes this paper.

\section{Problem Formulation} \label{sec:problem_formulation}
\noindent
Consider a dynamic web-services eco-system with $n$ users, denoted as $\mathcal{U} = \{u_1, u_2, \ldots, u_n\}$ and $m$ services, denoted as $\mathcal{S} = \{s_1, s_2, \ldots, s_m\}$. We monitor their interaction over the past ${\mathcal{T}}$ time-steps through a QoS invocation tensor ${\mathcal{Q}}_{n \times m \times {\mathcal{T}}}$ in terms of a QoS parameter $q$. It may be noted that ${\mathcal{Q}}$ is a sparse tensor where valid entries are positive real numbers, while zeros represent invalid entries.
The objective of the classical temporal QoS prediction is to estimate the missing value of ${\mathcal{Q}}(i, j, t)$, which indicates the QoS value of service $s_j$ for user $u_i$ at time-step $t$. 

The classical temporal QoS prediction problem often encounters three main research challenges, which can be categorized into six key research questions (RQ):

\textbf{\emph{(i) Data deficiency}}: Insufficient QoS observations lead to high sparsity in $\mathcal{Q}$ due to:
    
\----  
\emph{RQ1. Limited user-service interactions}: Users $u_i \in \mathcal{U}$ do not interact with all services $s_j \in \mathcal{S}$ at each time-step $t$, resulting in a sparse QoS vector ${\mathcal{Q}}^t (i,.)$, where ${\mathcal{Q}}^t$ denotes the QoS invocation matrix at $t^\text{th}$ time-step. How can we effectively leverage this sparse data for reliable QoS predictions?
   
\----   \emph{RQ2. Cold-start problem}: Newly added users or services have little to no interaction data. How can we effectively predict QoS for these new users or services despite the absence of historical data?
  
\textbf{\emph{(ii) Data credibility}}: Unreliable QoS data can degrade prediction accuracy, driven by:  
    
\----   \emph{RQ3. Outliers problem}: Dynamic factors such as limited network bandwidth or competition among service providers can cause outliers in QoS data. How can we effectively suppress these outliers to ensure reliable QoS predictions?
       
\----   \emph{RQ4. Greysheep problem}: Some users or services may exhibit a unique QoS invocation pattern that significantly deviates from others, termed as {\emph{greysheep}} \cite{GS}. How can we detect and leverage their characteristics for better predictions?

\textbf{\emph{(iii) Data representation problem}}: The QoS invocation pattern of a user or service can vary due to various known and unknown factors. Thus, capturing the diverse characteristics of users and services is crucial for accurate feature representation and QoS prediction: 

\----   \emph{RQ5. Domain-specific features}: How can we effectively use domain-specific features, e.g., user preferences and service characteristics, to improve prediction stability, speed, and accuracy?

\----   \emph{RQ6. Higher-order features}: In addition to domain knowledge features, it is crucial to explore complex, higher-order features such as spatial and temporal granularity for better data representation learning. How can we harness these intricate features for enhanced QoS prediction?

The objective of this paper is to introduce an anomaly-resilient temporal QoS prediction framework designed to address the aforementioned research questions. 
\section{Proposed Method}\label{sec:proposed_method}
\begin{figure*}[ht]
    \centering
    \includegraphics[width=\textwidth]{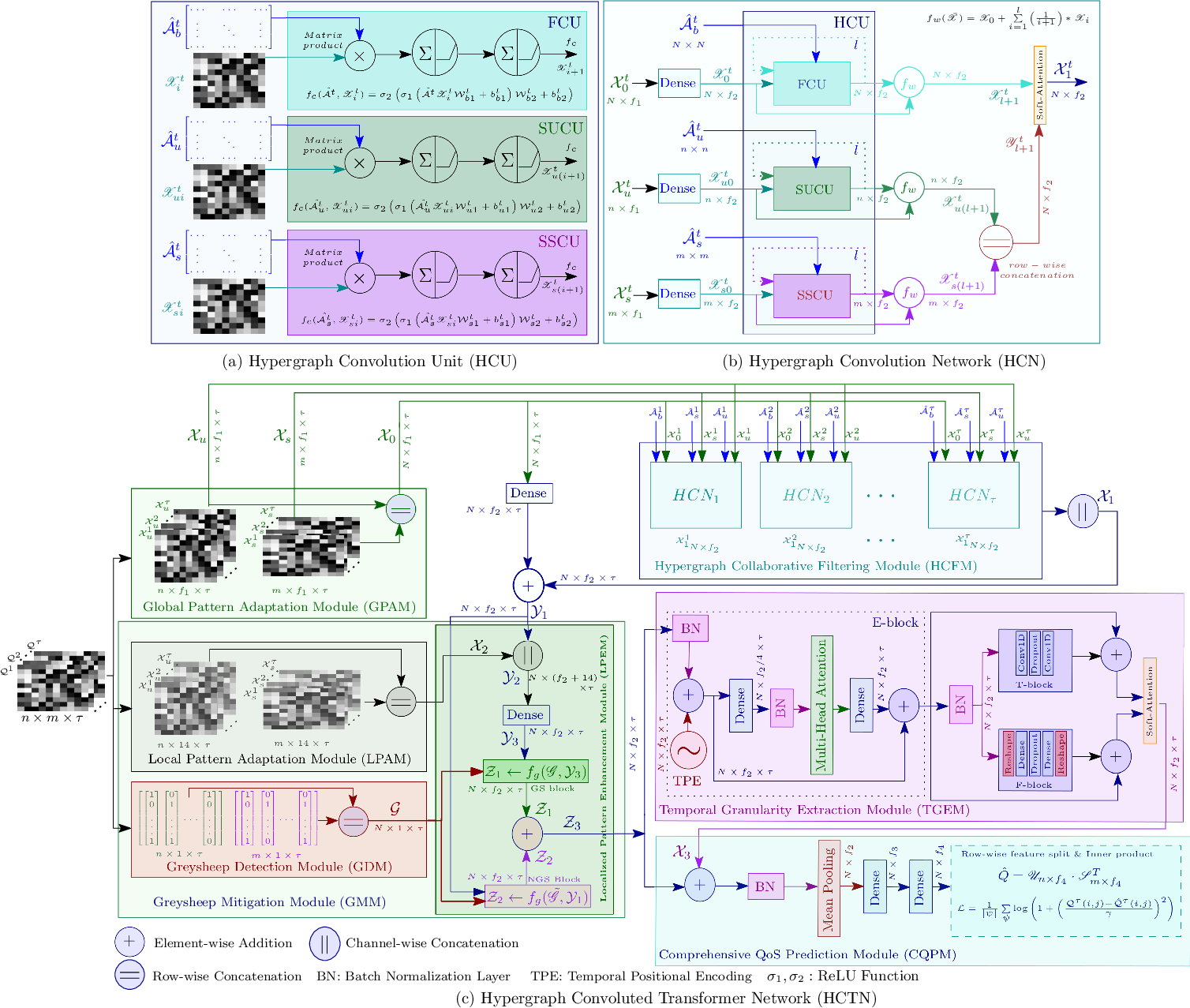}
    \caption{The architecture of HCTN: (a) Hypergraph Convolution Unit (HCU), (b) Hypergraph Convolution Network (HCN), (c) End-to-end architecture of Hypergraph Convoluted Transformer Network (HCTN) for real-time temporal QoS prediction}
    \label{fig:arch}
\end{figure*}

\noindent
This section introduces our proposed framework for anomaly-resilient real-time temporal QoS prediction comprising an end-to-end Hypergraph Convoluted Transformer Network (HCTN) that is composed of five primary modules: 
(\emph{A}) \emph{Global Pattern Adaptation Module (GPAM)}, 
(\emph{B}) \emph{Hypergraph Collaborative Filtering Module (HCFM)}, 
(\emph{C}) \emph{Greysheep Mitigation Module (GMM)} with two sub-modules: 
the Greysheep Detection Module (GDM), 
and the Local Pattern Adaptation Module (LPAM), 
(\emph{D}) \emph{Temporal Granularity Extraction Module (TGEM)}, and 
(\emph{E}) \emph{Comprehensive QoS Prediction Module (CQPM)}. 
By leveraging the diverse features obtained from various modules of HCTN, the overall goal of the framework is to deliver highly accurate real-time temporal QoS predictions while effectively addressing the challenges outlined in Section \ref{sec:problem_formulation}. A brief overview of HCTN is presented in Fig. \ref{fig:arch}. 
We now illustrate each module of HCTN in detail. 
%


\begin{figure}[!b]
    \centering
    \includegraphics[width=0.45\textwidth]{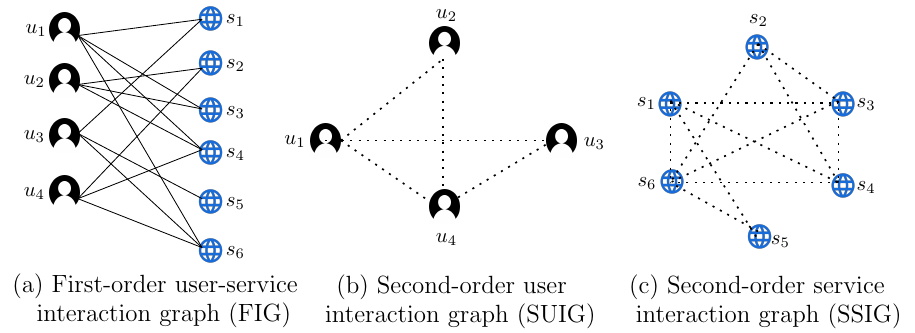}
    \caption{Decomposition of QIHG: (a) FIG, (b) SUIG, (c) SSIG}
    \label{fig:hypergraph}
\end{figure}

\subsection{Global Pattern Adaptation Module (GPAM)}
\noindent
The primary purpose of the GPAM is to provide domain-specific features by preserving the global characteristics of user-service interactions at a given time-step $t$. These features contribute to generating initial feature embeddings for each user and service. GPAM employs Non-negative Matrix Decomposition (NMD) \cite{NMF} to uncover hidden QoS invocation patterns among users and services.

Given a QoS invocation matrix ${\mathcal{Q}}^t$ for $n$ users and $m$ services, NMD decomposes ${\mathcal{Q}}^t$ into two low-rank latent features matrices, $\mathcal{X}_u^t$ and $\mathcal{X}_s^t$, such that: 
$${\small{
{\mathcal{Q}}^t_{n \times m} (i, j) \approx \left[(\mathcal{X}_u^t)_{n \times f_1} \cdot ({\mathcal{X}_s^t})_{m \times f_1}^T\right](i, j), ~~\forall \mathcal{Q}^t(i, j) \geq 0}}
$$
By applying NMD to each of the previous $\tau$ time-steps, GPAM extracts $f_1$-dimensional features for users and services, as depicted in Fig. \ref{fig:arch}(c). This process yields two feature tensors: 
$\mathcal{X}_u$ with dimensions $n \times f_1 \times \tau$, and
$\mathcal{X}_s$ with dimensions $m \times f_1 \times \tau$, corresponding to users and services, respectively. 
These user and service features are further combined into a user-service feature tensor $\mathcal{X}_0$ with dimensions $N \times f_1 \times \tau$, where $N = n + m$, by row-wise concatenating each $\mathcal{X}^t_u$ and $\mathcal{X}^t_s$.
These feature embeddings are utilized in the next module discussed in Section \ref{sec:hcfm}. Notably, GPAM produces dense feature tensors, which help mitigate the cold-start problem without requiring additional interventions.

\subsection{Hypergraph Collaborative Filtering Module (HCFM)} \label{sec:hcfm}
\noindent
GPAM effectively provides collaborative features that partially address the data-sparsity problem. However, it falls short in capturing higher-order, non-linear, and complex features of users and services that are essential for high prediction performance. To address this, various GCN-based frameworks \cite{QoSGNN,trqp,tpmcf,BGCL} have been introduced in the literature, utilizing bipartite graph representations of user-service QoS interactions. However, user-service interactions can extend beyond bipartite connections, leading to multidimensional relationships represented by hyper-edges. In this paper, we propose a Hypergraph Convolution Network (HCN), as presented in Fig. \ref{fig:arch}(b), to exploit these connections. HCN primarily leverages graph convolution over various graph representations constructed using different hyper-edges, as shown in Fig. \ref{fig:hypergraph}. Before discussing the details of this module, we first introduce the concept of a hypergraph in the context of QoS invocation.

\subsubsection{Hypergraph to Represent QoS Invocation}\label{sec:hypergraph} 
The connectivity between users and services to form a graph structure can be established through various methods, such as by shared autonomous systems or geographical locations. However, due to privacy concerns, we may not always have access to such details. To address this issue, we construct a hypergraph structure based on their past interactions recorded in the QoS invocation matrix at a given time-step $t$. For each time-step $t$, we define the QoS invocation hypergraph in Definition \ref{def:hypergraph}.

\begin{definition}{[\textbf{QoS Invocation Hypergraph (QIHG)}]}: \label{def:hypergraph}
  Given a set of users $\mathcal{U}$ and services $\mathcal{S}$, the QoS Invocation Hypergraph (QIHG) at time-step $t$, denoted by $\mathcal{G}^t$, is defined as $\mathcal{G}^t = (\mathcal{V}_u \cup \mathcal{V}_s, \mathcal{E}^t)$. Here, $\mathcal{V}_u$ represents the set of users $\mathcal{U}$, and $\mathcal{V}_s$ represents the set of services $\mathcal{S}$. The hyperedges $\mathcal{E}^t$ extend beyond simple pairwise connections and are of two types: (i) a hyperedge $\{v_i^u, v_j^s, v_k^u\}$, indicating that both users $u_i$ and $u_k$ corresponding to $v_i^u$ and $v_k^u \in \mathcal{V}_u$ invoked service $s_j$ corresponding to $v_j^s \in \mathcal{V}_s$ at time-step $t$; and (ii) a hyperedge $\{v^s_i, v^u_j, v^s_k\}$, indicating that both services $s_i$ and $s_k$ corresponding to $v_i^s$ and $v_k^s \in \mathcal{V}_s$ were invoked by user $u_j$ corresponding to $v_j^u \in \mathcal{V}_u$ at time-step $t$.
    \hfill$\blacksquare$
\end{definition}

\noindent
We simplify the complex hypergraph $\mathcal{G}^t$ by decomposing it into three distinct graphs (shown in Fig. \ref{fig:hypergraph}), each capturing a specific type of relationship among users and services:

\emph{(a) First-order user-service Interaction Graph (FIG)}: 
The FIG, denoted as ${\mathcal{G}}^t_f = (\mathcal{V}_u \cup \mathcal{V}_s, \mathcal{E}^t_f)$, captures direct interactions between users and services. An edge $(v_i^u, v_j^s) \in \mathcal{E}^t_f$ exists if the user corresponding to $v_i^u \in \mathcal{V}_u$ invoked the service corresponding to $v_j^s \in \mathcal{V}_s$ at time-step $t$.
$\mathcal{G}^t_f$ is represented by an adjacency matrix $\mathcal{A}^t \in \{0,1\}_{N \times N}$, where $N = n + m$.

\emph{(b) Second-order User Interaction Graph (SUIG)}: The SUIG, denoted by $\mathcal{G}^t_u = (\mathcal{V}_u, \mathcal{E}^t_u)$, represents the relationship between two users. An edge $(v_i^u, v_j^u) \in \mathcal{E}^t_u$ exists if the users corresponding to $v_i^u$ and $v_j^u$ both invoked at least one common service at $t$, i.e., $\exists s_k \in \mathcal{S}$, such that $\mathcal{Q}(i,k,t) \ne 0$ and $\mathcal{Q}(j,k,t) \ne 0$, which means, $\exists v^s_k \in \mathcal{V}_s$, such that $\{v_i^u, v_k^s, v_j^u\} \in \mathcal{E}^t$.

\emph{(c) Second-order Service Interaction Graph (SSIG)}: 
The SSIG, denoted by $\mathcal{G}^t_s = (\mathcal{V}_s, \mathcal{E}^t_s)$, represents the relationship between two services. An edge $(v_i^s, v_j^s) \in \mathcal{E}^t_s$ exists if the services corresponding to $v_i^s$ and $v_j^s$ both were invoked by at least one common user at $t$.
This means $\exists u_k \in \mathcal{U}$ such that $\mathcal{Q}(k,i,t) \ne 0$ and $\mathcal{Q}(k,j,t) \ne 0$, indicating that $\{v_i^s, v_k^u, v_j^s\} \in \mathcal{E}^t$ for some $\exists v^u_k \in \mathcal{V}_u$.

\subsubsection{Hypergraph Convolution Units (HCU)}\label{sec:h_conv}
We now explain the process of extracting higher-order collaborative features using hypergraph convolution on various graphs, including $\mathcal{G}^t_f$, $\mathcal{G}^t_u$, and $\mathcal{G}^t_s$, each representing distinct relationships. Fig. \ref{fig:arch}(a) provides an overview of the Hypergraph Convolutional Unit (HCU) architecture. 
The HCU consists of three convolution units: 
FCU, SUCU, and SSCU, as described below. 

\emph{(a) First-order Convolution Unit (${FCU}$):} At each time-step $t$, the FCU operates on the FIG $\mathcal{G}^t_f$. It takes two inputs: the normalized adjacency matrix $\hat{\mathcal{A}}^t$ of $\mathcal{G}^t_f$ and a pre-processed feature embedding $\mathscr{X}^t_i$ representing all users and services. The FCU then applies graph convolution \cite{GCN} function $f_c$ to these inputs to extract nonlinear collaborative features, as outlined in the equation presented in Fig. \ref{fig:arch}(a).
The normalized adjacency matrix $\hat{\mathcal{A}}^t = \left(\mathcal{D}^t \right)^{-1/2} \cdot \left(\mathcal{A}^t + \mathbb{I} \right ) \cdot \left(\mathcal{D}^t \right)^{-1/2}$ is obtained using symmetric normalization.
The diagonal degree matrix, derived from $\mathcal{A}^t$, and defined as $\mathcal{D}^t(i, i) = \sum_{j} \mathcal{A}^t({i,j})$, is used to calculate $\hat{\mathcal{A}}^t$. 
It may be noted that using $\hat{\mathcal{A}}^t$ helps mitigate numerical instability and sensitivity issues caused by high-degree nodes, unlike when using ${\mathcal{A}}^t$ directly. Additionally, the identity matrix ($\mathbb{I}$) is added during normalization to include self-features in the convolution process.

\emph{(b) Second-order User Convolution Unit (SUCU) and Second-order Service Convolution Unit (SSCU):} 
At each time-step $t$, the SUCU and SSCU operate on $\mathcal{G}^t_u$ and $\mathcal{G}^t_s$, respectively, applying graph convolution functions $f_c(\hat{\mathcal{A}}^t_u,~ \mathscr{X}^t_{ui})$ and $f_c(\hat{\mathcal{A}}^t_s,~ \mathscr{X}^t_{si})$ to independently extract nonlinear collaborative features $\mathscr{X}^t_{u(i+1)}$ for users and $\mathscr{X}^t_{s(i+1)}$ for services, as shown in the equations in Fig. \ref{fig:arch}(a).

Here, $\hat{\mathcal{A}}^t_u$ and $\hat{\mathcal{A}}^t_s$ represent the normalized matrices that the SUCU and SSCU receive as input. However, unlike the FCU, which uses adjacency matrices derived from $\mathcal{G}^t_f$, we instead leverage the incidence matrix $\mathcal{H}^t$, which is also derived from $\mathcal{G}^t_f$. This matrix is defined as $\mathcal{H}^t = (h^t_{ij})_{n \times m} \in \{0,1\}_{n \times m}$, where $h^t_{ij} = 1$ if $\mathcal{Q}(i, j, t) \ne 0$. The derivation to compute $\hat{\mathcal{A}}^t_u$ and $\hat{\mathcal{A}}^t_s$ is presented in Eq. \ref{eq:adj_scu_u}.  
%
\begin{equation} \label{eq:adj_scu_u}
\small 
\begin{split}       
    \hat{\mathcal{A}^t_u} &= \left(\mathcal{D}^t_u\right)^{-1/2} \cdot \mathcal{H}^t \cdot \left(\mathcal{D}^t_s\right)^{-1} \cdot \left(\mathcal{H}^t\right)^{T} \cdot \left(\mathcal{D}^t_u\right)^{-1/2} ;\\
     \hat{\mathcal{A}^t_s} &= \left(\mathcal{D}^t_s\right)^{-1/2} \cdot \left(\mathcal{H}^t\right)^T \cdot \left(\mathcal{D}^t_u\right)^{-1} \cdot \mathcal{H}^t \cdot \left(\mathcal{D}^t_s\right)^{-1/2} 
\end{split} 
\end{equation}
where, the degree matrices $(\mathcal{D}^t_u)_{n \times n}$ and $(\mathcal{D}^t_s)_{m \times m}$ are derived from the adjacency matrices ${\mathcal{A}}^t_u$ and ${\mathcal{A}}^t_s$, corresponding to $\mathcal{G}^t_u$ and $\mathcal{G}^t_s$, respectively. These degree matrices are computed in a similar manner to $\mathcal{D}^t$.

We now introduce the Hypergraph Convolution Network (HCN), where the HCU serves as the primary component.

\subsubsection{Hypergraph Convolution Network (HCN)}
Fig. \ref{fig:arch}(b) illustrates an overview of the HCN. 
At each time-step $t$, HCN receives feature embeddings $\mathcal{X}^t_0$, $\mathcal{X}^t_u$, and $\mathcal{X}^t_s$ as inputs, which are generated by the GPAM module. 
These embeddings are first processed through a dense layer, yielding $\mathscr{X}^t_0$, $\mathscr{X}^t_{u0}$, and $\mathscr{X}^t_{s0}$, which are then fed into HCU in a loop $l$ times.

To counteract potential information loss caused by the over-smoothing issue commonly encountered with multiple graph convolution layers \cite{survey_gnn}, we aggregate the outputs of each layer using the function $f_w(\bar{\mathscr{X}}) = \mathscr{X}_0 + \sum_{i=1}^{l} \left({1}/{(i+1)}\right)  \mathscr{X}_i$, where $\bar{\mathscr{X}} = (\mathscr{X}_0, \mathscr{X}_1, \ldots, \mathscr{X}_l)$ is a tuple of the convolution outputs of length $(l + 1)$.
This method balances complexity and feature retention by assigning progressively smaller weights to deeper layer outputs, preserving critical information. 
Finally, we compute $f_w(\bar{\mathscr{X}^t})$, $f_w(\bar{\mathscr{X}^t_{u}})$, and $f_w(\bar{\mathscr{X}^t_{s}})$ to generate $\mathscr{X}^t_{l + 1}$, $\mathscr{X}^t_{u(l + 1)}$, and $\mathscr{X}^t_{s(l + 1)}$, respectively.

We then aggregate $\mathscr{X}^t_{u(l + 1)}$ and $\mathscr{X}^t_{s(l + 1)}$ through row-wise concatenation to form $\mathscr{Y}^t_{l + 1}$.
This concatenated matrix is then combined with $\mathscr{X}^t_{l + 1}$ using soft-attention \cite{soft_attn} to generate $\mathcal{X}^t_1$, which represents the refined, enriched collaborative features as the output of the HCN block.

\subsubsection{Hypergraph Collaborative Filtering} 
We now discuss the working of the HCFM. The HCFM leverages the HCN across the past $\tau$ time-steps. For each $HCN_t$ where $1 \le t \le \tau$, the inputs include the normalized matrices $\hat{\mathcal{A}}^t$, $\hat{\mathcal{A}}^t_u$, and $\hat{\mathcal{A}}^t_s$, along with the feature embeddings $\mathcal{X}^t_0$, $\mathcal{X}^t_u$, and $\mathcal{X}^t_s$. The outputs from each HCN block are concatenated channel-wise to produce the final embedding $\mathcal{X}_1$ of size $N \times f_2 \times \tau$.

It is worth noting that the HCFM leverages higher-order, nonlinear, and complex collaborative features, which help alleviate data sparsity issues by utilizing the additional interactions captured through $\mathcal{G}^t_u$ and $\mathcal{G}^t_s$, for each $t$, where $1 \le t \le \tau$. However, during the generation of these collaborative features, crucial initial features that are beneficial for cold-start scenarios may not be preserved. To address this, we introduce a skip connection from the GPAM, where $\mathcal{X}_0$ is resized and added to $\mathcal{X}_1$, producing the feature tensor $\mathcal{Y}_1$ of the same size. This skip connection mitigates the cold-start issue and provides stability in training the HCTN. In the next section, we incorporate the module designed to tackle the greysheep problem.

\subsection{Greysheep Mitigation Module (GMM)}
\noindent
While exploring collaborative features, GPAM and HCFM assume user and service preferences are consistent and QoS data remain reliable with time. However, a few user or service characteristics may not be coherent with others due to their deviant pattern of QoS invocations, referred to as Greysheep \cite{GS}. Finding such users (services) and selectively fueling them with their own features obtained from their QoS profiles is essential for robust QoS prediction. In this section, we specifically deal with the greysheep instances whose presence suffers collaborative filtering methods and degrades QoS prediction performance. 
This module consists of three key components:
(a) \emph{Greysheep Detection Module (GDM)}: 
Identifies {greysheep users (GU) and greysheep services (GS)} within the given QoS data. 
(b) \emph{Local Pattern Adaptation Module (LPAM)}: Extracts local features for users and services.
(c) \emph{Localized Pattern Enhancement Module (LPEM)}: Injects these local features to address the specific challenges posed by greysheep instances.

\subsubsection{Greysheep Detection Module (GDM)} 
\label{subsubsec:GDM} 
The primary objective of the GDM is to identify greysheep instances within the dataset. These instances refer to {GU or GS} whose preferences deviate significantly from the norm, resulting in lower QoS prediction performance. Inspired by the approach in \cite{GS} for detecting atypical instances, the GDM, as shown in Fig. \ref{fig:arch}(c), is specifically designed to identify these greysheep instances. Given the QoS invocation tensor $\mathcal{Q}_{n \times m \times \tau}$ for the past $\tau$ time-steps, the GDM employs a metric called the Greysheep Discrepancy Index (GDI) to assess each user and service for potential greysheep characteristics.


\emph{(a) Greysheep Discrepancy Index (GDI):} 
We adopt the concept introduced in \cite{GS} to define the GDI. 
The GDI of a user $u_i$ (denoted by $\mathbb{G}^{t}(u_i)$) or a service $s_j$ (denoted by $\mathbb{G}^{t}(s_j)$) is a scalar value derived by comparing the QoS invocation profile of $u_i$ (i.e., $\mathcal{Q}^t(i, .)$) or of $s_j$ (i.e., $\mathcal{Q}^t(., j)$) against the overall QoS mean for $u_i$ (represented by $\mu^t (u_i) = mean(\mathcal{Q}^t(i, .))$) or $s_j$  (represented by $\mu^t (s_j) = mean(\mathcal{Q}^t(., j))$). This comparison also takes into account the individual service or user means when $u_i$ interacts with a specific service or when $s_j$ is invoked by a particular user. Appendix A outlines the detailed GDI calculation. A large GDI value indicates the abnormality of $u_i$ or $s_j$, guiding the labeling of 
{GUs or GSs}. 

\emph{(b) Greysheep Labeling:} We define the greysheep indicator tensor $\mathcal{G}$, with dimensions $N \times 1 \times \tau$, as follows:
\begin{equation}\label{eq:git}\scriptsize
    \mathcal{G} (i, 1, t) =
    \begin{cases}
        1 & \text{ if } i \le n \text{ and } \mathbb{G}^t (u_i) >  \mu^t_{gu} + c_1 * \sigma^t_{gu}\\
        1 & \text{ if } i > n \text{ and } \mathbb{G}^t (s_i) >  \mu^t_{gs} + c_2 * \sigma^t_{gs}\\
        0 & \text{ otherwise }
    \end{cases}
\end{equation}
where, $\mu^t_{gu}$ and $\sigma^t_{gu}$ represent the mean and standard deviation of the GDI values for users, while $\mu^t_{gs}$ and $\sigma^t_{gs}$ represent the mean and standard deviation of the GDI values for services at time-step $t$. The positive constants $c_1$ and $c_2$ are two tunable hyperparameters.
This tensor is then utilized in subsequent operations to obtain fine-tuned features for both users and services.
In the next subsection, we introduce our next module, LPAM, which extracts user- and service-specific features that are particularly beneficial for greysheep instances, aiming to enhance prediction accuracy.

\subsubsection{Local Pattern Adaptation Module (LPAM)}
The objective of this module is to extract local patterns from the QoS invocation profiles of each user and service. For each time-step $t$ within the $\tau$-time-steps, the LPAM is used to obtain local features for each user $u_i \in \mathcal{U}$ and service $s_j \in \mathcal{S}$ based on their QoS profiles $\mathcal{Q}^t(i, ., t)$ and $\mathcal{Q}^t(., j, t)$. These local features consist of 14 basic statistical metrics, including minimum, maximum, mean, median, standard deviation, skewness, kurtosis, interquartile range, mean absolute deviation, median absolute deviation, root mean square, absolute energy, entropy, and peak-to-peak difference \cite{tsfel2020}. These features provide diverse quantitative insights into the central tendencies, variability, and distributions of the data. We store these features in a tensor $\mathcal{X}_2$ with dimensions $N{\times}14{\times}\tau$. In this tensor, the first $n$ entries of the first dimension correspond to users, while the remaining $m$ entries correspond to services. We then leverage $\mathcal{X}_2$ to manage GU and GS. In the next module, we will explain the process of injecting these features into the greysheep instances.

\subsubsection{Localized Pattern Enhancement Module (LPEM)} 
Since collaborative features alone are insufficient to fully represent the GU and GS, we enhance them with profile-specific features to support representation learning and improve QoS prediction performance. To achieve this, we concatenate the collaborative features, $\mathcal{Y}_1$, with profile-specific features, $\mathcal{X}_2$, along the second dimension of the tensor, resulting in $\mathcal{Y}_2$ with dimensions 
$N{\times}(f_2{+}14){\times}\tau$. $\mathcal{Y}_2$ is then processed through a fully connected layer to reshape the tensor, generating the output tensor $\mathcal{Y}_3$ with dimensions $N{\times}f_2{\times}\tau$.

To obtain the embedding for greysheep instances, we introduce a function $f_g$, as described in Eq. \ref{eq:gs_injection2}, which takes the greysheep indicator tensor $\mathcal{G}$ and $\mathcal{Y}_3$ to filter out features unique to greysheep instances from regular users and services. This results in the embedding $\mathcal{Z}_1 = f_g(\mathcal{G}, \mathcal{Y}_3)$.

For non-greysheep users and services, indicated by $\tilde{\mathcal{G}} = \left({\mathbb{1}}_{N \times 1 \times \tau} - \mathcal{G}\right)$, only $\mathcal{Y}_1$ is used, leading to the embedding $\mathcal{Z}_2 = f_g(\tilde{\mathcal{G}}, \mathcal{Y}_1)$. Finally, $\mathcal{Z}_1$ and $\mathcal{Z}_2$ are combined to form $\mathcal{Z}_3$, which retains the distinguishing features of both greysheep and non-greysheep instances. This $\mathcal{Z}_3$ is subsequently used in the next module.
\begin{equation} \label{eq:gs_injection2}
\scriptsize     
    f_g(\mathcal{G}, \mathcal{Y}) = [\mathcal{Z}^1 ~||~ \mathcal{Z}^2 ~||~ \ldots ~||~ \mathcal{Z}^\tau]; \quad
    \forall t = [1 .. \tau], \mathcal{Z}^t = (\mathcal{G}^t e) \odot \mathcal{Y}^t 
\end{equation}
where, $e$ is a $1 \times f_2$ sized vector of $1$s, $\odot$ denotes the 
Hadamard product, and $||$ represents channel-wise concatenation.

\subsection{Temporal Granularity Extraction Module (TGEM)}
\noindent
The features obtained from the previous modules are limited in their ability to capture the intricate temporal dynamics necessary for accurate temporal QoS prediction. To address this, we introduce the module TGEM, designed to capture temporal features across different scales. TGEM consists of three primary components:
(a) \emph{E-block}: A transformer encoder block \cite{TRANSFORMER} that includes multi-head attention ($MHA$) mechanisms to capture long-range dependencies.
(b) \emph{T-block}: A block dedicated to capturing fine-grained temporal features, which is composed of two 1D convolutional layers for localized temporal feature extraction.
(c) \emph{F-block}: A block for extracting fine-grained features at each individual time-step, consisting of two fully connected dense layers.
These components, as illustrated in Fig. \ref{fig:arch}(c), work together to capture temporal features at varying levels of granularity. 

\subsubsection{Temporal Dependency Extractor Block (E-block)}
This block is equipped with a transformer encoder \cite{TRANSFORMER}, incorporating $MHA$ mechanisms. The E-block applies multiple attention heads in parallel, followed by a fully connected dense layer to capture multiple contexts from the input embedding. Additionally, it incorporates positional encoding to overcome the limitation of $MHA$ in preserving the sequential information of the input data. We adopt sinusoidal positional encoding, similar to \cite{TRANSFORMER}, to maintain temporal sequence information.

Upon receiving the input $\mathcal{Z}_3$ from GMM, batch normalization ($BN$) is applied to re-center and re-scale the features, reducing internal covariate shift and improving training stability and speed \cite{batch_norm}. After calculating the temporal positional encoding (TPE) for each user-service feature at each time-step, we add the TPE to $BN(\mathcal{Z}_3)$ to produce the intermediate result $\mathcal{Z}_4$, which is then passed to the $MHA$ block. 
However, the classical transformer encoder faces scalability challenges with high-dimensional feature embeddings. To address this, rather than passing the combined results directly to the $MHA$ block, we introduce a fully connected dense layer, followed by another batch normalization step. This reduces the dimensionality of the input features by one-fourth, resulting in $\mathcal{Z}_5$ with dimension $N \times (f_2/4) \times \tau$.

\emph{Multi-Head Attention (MHA)}: 
$\mathcal{Z}_5$ contains the features for all users and services across all $\tau$ time-steps. For each user or service instance $i$ (out of the $N$ instances), the features across the $\tau$ time-steps, denoted as $\mathcal{Z}^{[i]}_5$, are rearranged and concatenated row-wise to form the feature matrix $\mathcal{F}_i$, with dimensions $\tau \times (f_2/4)$. This matrix is then used as input for the $MHA$ that operates in three steps as follows: First, for each $j^{\text{th}}$ head, we generate three different representations of $\mathcal{F}_i$, namely, the Query $\mathbb{Q}_j = \mathcal{F}_i \mathcal{W}^{1[i]}_j$, the Key $\mathbb{K}_j = \mathcal{F}_i \mathcal{W}^{2[i]}_j$, and the Value $\mathbb{V}_j = \mathcal{F}_i \mathcal{W}^{3[i]}_j$, where $\mathcal{W}^{k[i]}_j$ (for $k \in {1, 2, 3}$) are learnable weights. For simplicity, we refrain from writing bias terms in this representation.
In the second step, we compute the scaled-dot product attention using $\mathbb{Q}^{[i]}_j$, $\mathbb{K}^{[i]}_j$, and $\mathbb{V}^{[i]}_j$ for each $j^{\text{th}}$ head, as shown in Eq. \ref{eq:sdpa}, where $d_k$ represents the dimension of the key used to scale dot-product. 
$\mathbb{H}^{[i]}_j$ accommodates self-attentive output embedding for each head $j$. 
\begin{equation}
    \label{eq:sdpa} \small
    \mathbb{H}^{[i]}_j(\mathbb{Q}^{[i]}_j, \mathbb{K}^{[i]}_j, \mathbb{V}^{[i]}_j) = softmax \left(({\mathbb{Q}^{[i]}_j \cdot (\mathbb{K}^{[i]}_j)^T})/{\sqrt{d_k}}\right) \mathbb{V}^{[i]}_j
\end{equation}
We employ total $h_n$ numbers of heads in E-block.
In the third and final step, the output embeddings from all heads are concatenated, followed by a linear transformation to produce the output $\mathcal{Z}^{[i]}_6 = MHA (\mathbb{Q}^{[i]}, \mathbb{K}^{[i]}, \mathbb{V}^{[i]}) = \left(\mathbb{H}^{[i]}_1 ~||~ \mathbb{H}^{[i]}_2 ~||~ \ldots ~||~ \mathbb{H}^{[i]}_{h_n}\right) \mathcal{W}$;  
where, $\mathcal{W}$ is learnable weight.
Each head captures a specific temporal feature, and their combination captures the different contexts within the input embedding. Finally, $\mathcal{Z}^{[i]}_6$ is reshaped and combined across all user-service instances to form $\mathcal{Z}_6$.
After the $MHA$ operation, $\mathcal{Z}_6$ is passed through a dense layer to restore its original dimensions, $N \times f_2 \times \tau$. This restored tensor is then added to $\mathcal{Z}_5$ using a residual connection, which preserves the collaborative characteristics for users and services, resulting in $\mathcal{Z}_7$ (refer to E-block of TGEM in Fig. \ref{fig:arch}(c)). 
$\mathcal{Z}_7$ is then passed through batch normalization, producing $\mathcal{Z}_8$ that is subsequently fed into the T-block and F-block. 
These blocks are explained next to obtain multi-granularity features from 
$\mathcal{Z}_8$.

\subsubsection{Fine-Grained Temporal Analyzer Block (T-block)}
To capture fine-grained temporal features, we employ the T-block, which utilizes two 1D convolutional layers (Conv1D) with a dropout layer \cite{dropout} between them, applied along the temporal dimension of $\mathcal{Z}_8$. The first Conv1D layer, with $\tau / 4$ filters, performs temporal pooling, reducing the temporal dimension by a factor of four, resulting in an output of $N \times f_2 \times (\tau/4)$. This step ensures that fine-grained features are effectively extracted. The second Conv1D layer, with $\tau$ filters, restores the temporal dimension to match that of $\mathcal{Z}_8$. Finally, we apply a residual connection by adding $\mathcal{Z}_8$ to the output, forming the final output $\mathcal{Z}_T$.

\subsubsection{Time-Step Feature Refiner Block (F-block)} 
Parallel to the T-block, the F-block is designed to capture coarse-grained features. It employs two fully connected dense layers applied over the feature dimension (i.e., the second dimension of the three-dimensional tensor) of the input, with a dropout layer placed between them. In the F-block, the input tensor $\mathcal{Z}_8$ is reshaped, interchanging the temporal and feature dimensions. The first dense layer applies a nonlinear transformation that reduces the feature dimension to one-fourth, while the second dense layer restores it to its original dimension through another nonlinear transformation. The output from the second layer is reshaped back and added to $\mathcal{Z}_8$ via a residual connection from the E-block, resulting in the final output embedding, $\mathcal{Z}_F$.

Finally, we apply soft-attention over $\mathcal{Z}_T$ and $\mathcal{Z}_F$ to refine the temporal dynamics, yielding $\mathcal{X}_3$.
It is important to note that while the parallel T-block and F-block enable multi-granularity feature extraction, the E-block captures various meaningful contexts. The output of the TGEM, $\mathcal{X}_3$, is then passed to the comprehensive QoS prediction module for final prediction.

\subsection{Comprehensive QoS Prediction Module (CQPM)}
\noindent
The CQPM predicts the QoS of a user–service pair at time-step $\tau$. It fuses the feature embeddings $\mathcal{Z}_3$ (from GMM) and $\mathcal{X}_3$ (from TGEM) via feature addition followed by batch normalization. Mean pooling aggregates historical features across the previous $\tau$ steps, and the result is passed through two fully connected layers. The final layer output is partitioned row-wise into user and service feature matrices, $\mathscr{U}_{n\times f_4}$ and $\mathscr{S}_{m\times f_4}$. The QoS estimate is then obtained via their inner product: $\hat{\mathcal{Q}}^{\tau} = \mathscr{U}_{n\times f_4} \cdot \mathscr{S}^{T}_{m\times f_4}$.


\subsection{Training and Prediction}
\noindent
To address the issue of outliers, we adopt two strategies. First, we adopt the Cauchy loss, denoted by $ {\mathcal{L}}$, as the training objective. The Cauchy loss, as shown in Eq. \ref{eq:loss_function}, is more robust to outliers compared to standard error metrics.
\begin{equation}\label{eq:loss_function}\scriptsize
 {\mathcal{L}} = \frac{1}{|\psi|} \sum_{{\mathcal{Q}}^{\tau}(i, j) \in \psi} \log\left(1 + \left(\left({{\mathcal{Q}}^{\tau}(i, j) - \hat{\mathcal{Q}}^{\tau}(i, j)}\right)/{\gamma}\right)^2\right)
\end{equation}
where, $\psi$ represents the set of valid entries in ${\mathcal{Q}}^{\tau}$, and $\gamma$ is a scale hyper-parameter.
Second, we apply the unsupervised Isolation Forest algorithm \cite{isolation-forest} to detect and remove a fixed percentage of outliers, controlled by the hyperparameter $\lambda$, highlighting HCTN’s performance without outlier interference.
AdamW optimizer \cite{adamw} is used here to train our model.
\section{Experiments}\label{sec:experiments}
\noindent
Our framework was trained offline using TensorFlow 2.16.1 with Python 3.10.13, leveraging an NVIDIA GeForce RTX 4080 GPU. For performance evaluation, the model was tested on a system equipped with an AMD Ryzen-9 7950X 16-core processor and 32 GB RAM.

\begin{table}[!b]
    \centering
    \scriptsize
        \centering
        \caption{WSDREAM-2 \cite{WSDREAM} dataset specifications}
        \begin{tabular}{l| c c}
        \hline
         Dataset Statistics & RT (seconds) & TP (kbps) \\ \hline 
         \# $\mathcal{U}$, \# $\mathcal{S}$, \# $\mathcal{T}$ & 142, 4500, 64 & 142, 4500, 64   \\ 
         \# Records, Density & 30170567, 73.77\% & 25652011, 62.72\% \\ 
        Range & (0.001, 19.999) & (3.65e-5, 6726.834)\\ 
        Mean, Std. Dev. & 3.1773, 6.1279 & 11.3449, 54.2759 \\ \hline
        \end{tabular}
        \label{tab:dataset1}
\end{table}

\begin{table}[!b]
    \centering
    \scriptsize
    \caption{Training-testing splits}
        \begin{tabular}{c| c| c c c c c}
            \hline
            Training Density & $\psi ~(\%)$ & 5 & 10 & 15 & 20 & 50 \\ \hline
            \multirow{2}{*}{QoS parameters} & RT (D1) & D1.1 & D1.2 & D1.3 & D1.4 & D1.5 \\
            & TP (D2) & D2.1 & D2.2 & D2.3 & D2.4 & D2.5 \\ \hline
        \end{tabular}
        \label{tab:dataset2}
\end{table}

\begin{table*}
    \centering
     \scriptsize 
        \begin{minipage}{0.70\textwidth}
        \captionof{table}{Performance comparison of HCTN with previous methods}
        \adjustbox{max width=\linewidth}{
        \begin{tabular}{l | c | c c c c c | c c c c c}
        \hline 
        \multirow{2}{*}{Methods} & Loss & \multicolumn{5}{c|}{MAE} & \multicolumn{5}{c}{RMSE} \\ \cline{3-7} \cline{8-12}
        & Function & D1.1 & D1.2 & D1.3 & D1.4 & D1.5 & D1.1 & D1.2 & D1.3 & D1.4 & D1.5 \\ \hline 
        WSPred \cite{WSPRED} & MSE & 2.5580  & 2.4990 & 2.4100 & 2.3000 & 2.1266 & 4.3626 & 4.2892 & 4.2000 & 4.1500 & 3.8943   \\ 
        TUIPCC \cite{TUIPCC} & - & 2.4088 & 2.2118 & 2.0910 & 1.9531 & 1.2636 & 4.8319 & 4.7751 & 4.6329 & 4.4454 & 3.1722 \\
        GFEN \cite{GFEN} & MSE & 2.3168 & 2.2901 & 2.2614 & 2.2968 & 1.9395 & 5.5295 & 5.4669 & 5.3925 & 5.1024 & 4.9139 \\
        SCATSF \cite{SCATSF} & Huber & 2.2657 & 2.2203 & 2.1234 & 2.0273 & 1.6352 & 5.4799 & 5.2403 & 4.9461 & 4.9326 & 3.9418 \\

        NNCP \cite{NNCP} &  MSE & 1.8000 & 1.7500 & 1.5500	& 1.5800 & 1.3000 & 3.9000 & 4.0500 & 3.7500 & 3.5000 & 2.9000  \\
        TRCF \cite{TRCF} & - & 1.7574 & 1.3319 & 1.1229 & 1.0132 & 0.8714 & 4.7187 & 3.9072 & 3.4028 & 3.0563 & 2.3592 \\
        DeepTSQP \cite{DeepTSQP} & MSE & 1.7329 & 1.5241 & 1.4540 & 1.3137 & 0.9710 & 4.1902 & 3.8777 & 3.5964 & 3.1313 & 2.3576 \\
        PLMF \cite{PLMF} & MSE & 1.4338 & 1.3031 & 1.2439 & 1.2209 & 1.1533 & 2.9201 & 2.7817  & 2.7185 & 2.6946 & 2.6278 \\

         STGCN \cite{stgcn} &  MSE &  1.3995 &  1.3088 &  1.2642 &  1.2308	&  1.1956 &  3.5544 &  3.4332 &  3.3461 &  3.3011 &  3.2512 \\

        RNCF \cite{RNCF} & MSE & 1.3386 & 1.2713 & 1.2069 & 1.1583 & 1.0762 & 3.1369 & 2.8542 & 2.7789 & 2.6038 & 2.6038 \\
        TaTruSR \cite{TaTruSR} & - & 1.3029 & 1.1195 & 1.0644 & 1.0253 & 0.9462 & 3.4266 & 2.9562 & 2.7738 & 2.6927 & 2.5351 \\
        CTF \cite{CTF} & Cauchy & 1.2884 & 1.1753 & 1.1504 & 1.0947 & 1.0430 & 2.9253 & 2.7172 & 2.6368 & 2.5769 & 2.4797 \\
        BNLFT \cite{BNLFT} & MSE & 1.2600 & 1.2400 & 1.2300	& 1.2200 & 1.2000 & \underline{2.8000} & \underline{2.5000} & \underline{2.4800} & 2.4500 & 2.4000  \\ 
        GMCL \cite{GMCL} & MAE + CL & 1.1572 & 1.1058 & 1.0748 & 1.0294 & 0.9542 & 3.3473 & 3.3285 & 3.1860 & 3.0952 & 2.9768  \\

         STGFT \cite{stgft} & 
         MAE & 
         1.1098 &
         0.8812 &
         0.7265 &
         0.6512 &
         0.4536 &
         3.1857 &
         2.9613 &
         2.6520 &
         2.4008 &
         2.2116 \\ 
          
        TPMCF \cite{tpmcf} & Cauchy & \underline{0.9735} & \underline{0.8132} & \underline{0.7168} & \underline{0.6159} & \underline{0.2528} & 2.9648 & 2.6856 & 2.6191 & \underline{2.2927} & \underline{1.5832} \\ \hline
        \textbf{HCTN} & \textbf{Cauchy} & \textbf{0.8453} & \textbf{0.6765} & \textbf{0.6133} & \textbf{0.5256} & \textbf{0.2237} & \textbf{2.4486} & \textbf{2.2336} & \textbf{2.1074} & \textbf{1.9754} & \textbf{1.2853} \\ \hline
        
        \multicolumn{2}{c|}{$\mathcal{I}$ (\%)} & 13.17\% & 16.81\% & 14.44\% & 14.66\% & 11.51\% & 12.55\% & 10.66\% & 15.02\% & 13.84\% & 18.82\% \\
        \hline \hline
        
        &  & D2.1 & D2.2 & D2.3 & D2.4 & D2.5 & D2.1 & D2.2 & D2.3 & D2.4 & D2.5 \\ \hline

        TUIPCC \cite{TUIPCC} & - & 15.2948 & 14.9539 & 14.0606 & 13.5154 & 8.8786 & 44.1169 & 43.8122 & 43.7239 & 43.1728 & 33.2959 \\
        
        GFEN \cite{GFEN} & MSE & 13.8755 & 13.8538 & 13.7270 & 14.3370 & 12.9108 & 60.3329 & 58.4838 & 60.2614 & 60.1972 & 59.4645 \\
        
        SCATSF \cite{SCATSF} & Huber & 13.1707 & 12.3124 & 11.9334 & 11.0718 & 9.2442 & 59.8496 & 58.5441 & 57.7450 & 56.4208 & 52.4242 \\

        TaTruSR \cite{TaTruSR} & - & 11.1396 & 9.0643 & 8.2291 & 7.6819 & 6.6458 & 52.1592 & 42.1581 & 39.7379 & 38.0506 & 33.0270 \\

        DeepTSQP \cite{DeepTSQP} & MSE & 9.6616 & 8.3258 & 8.8682 & 9.3145 & 10.4283 & 49.4513 & 43.1617 & 41.1476 & 38.9401 & 36.3836 \\

        PLMF \cite{PLMF} & MSE & 9.2414 & 9.2217 & 9.1666 & 9.1091 & 9.1751 & 51.7130 & 51.6902 & 51.3862 & 51.1958 &  51.8412 \\
        
        TRCF \cite{TRCF} & - & 9.1978 & 6.4491 & 4.8417 & 4.0694 & 3.1343 & 47.8063 & 38.8540 & 31.0267 & 28.0598 & 20.5362 \\
        
        WSPred \cite{WSPRED} & MSE & 8.2761 & 8.0131 & 7.8500 & 7.6000 & 6.8000 & 39.0962 & 38.6251 & 38.5000 & 37.6000 & 36.5724  \\

        TPMCF \cite{tpmcf} & Cauchy & 7.0860 & 5.7447 & 4.8590 & 4.5136 & \underline{1.4232} & 40.1241 & 37.6999 & 34.7010 & 30.5240 & 20.7425 \\ 

        RNCF \cite{RNCF} & MSE & 6.4916 & 5.6454 & 5.2319 & 5.4083 & 4.5919 & 32.7867 & 28.4236 & 26.8409 & 25.3949 & 22.8516 \\

         STGCN \cite{stgcn} &  MSE &  5.5987 &  4.9839 &  5.0139 &  4.5411 &  3.9105 & 
         46.9290 &  44.9156 &  42.7702 &  42.5804 &  38.0443 \\

        GMCL \cite{GMCL} & MAE + CL & 5.3213  & 4.4768 & 4.4021 & 4.1963 & 4.1603 & 36.9408 & 31.1152 & 30.2974 & 27.7784 & 26.7878  \\ 
        
        NNCP \cite{NNCP}   & MSE & 5.1000	& 4.5000 & 4.2500 & 4.1000 & 3.7500 & 32.5000 & 32.5000 & 25.5000 & 24.5000 & 23.5000 \\ 

         STGFT \cite{stgft} &  MAE & 
         4.3302 &
         4.2174 & 
         3.9836 & 
         3.7692 &
         3.2256 &
         28.5001 
        &  26.4642 
        &  24.7985 &
         23.2439 &
         22.1006
         \\ 

        BNLFT \cite{BNLFT} & MSE & 4.2500 & 4.1000 & 3.9000 & 3.7500 & 3.4000 & \underline{24.5000}	& \underline{23.4000} & \underline{21.5000} & \underline{22.6000} & \underline{19.9000} \\

        CTF \cite{CTF} & Cauchy & \underline{4.0279}  & \underline{3.6458} & \underline{3.5602} & \underline{3.5578} & 3.4856 & 32.0197 & 31.6197 & 31.2808 & 31.4697 & 28.6353 \\ \hline

         \textbf{HCTN} & \textbf{Cauchy} & \textbf{3.9693} & \textbf{3.5486} & \textbf{3.3071} & \textbf{2.8671} & \textbf{0.7404} & \textbf{23.8219} & \textbf{21.3090} & \textbf{20.8300} & \textbf{18.8000} & \textbf{4.8526} \\ \hline
 
         \multicolumn{2}{c|}{$\mathcal{I}$ (\%)} & 1.45\% & 2.67\% & 7.11\% & 19.41\% & 47.98\% & 2.77\% & 8.94\% & 3.12\% & 16.81\% & 75.62\% \\ \hline
         \multicolumn{12}{r}{CL: Contrastive Loss; \quad The performance of the second-best method is \underline{underlined}}
        \end{tabular}}
        \label{tab:soa_table}
    \end{minipage}
    \hfill
\begin{minipage}{0.29\textwidth}
    \centering
     \vspace{0.25cm}
        \begin{tikzpicture}
            \begin{axis}[
                width=\textwidth,
                height=0.6\textwidth,
                ylabel={Time (second)},
                ybar=5pt,
                symbolic x coords={TUIPCC \cite{TUIPCC}, BNLFT \cite{BNLFT}, WSPred \cite{WSPRED}, CTF \cite{CTF}, NNCP \cite{NNCP},SCATSF \cite{SCATSF}, TPMCF \cite{tpmcf}, GFEN \cite{GFEN}, DeepTSQP \cite{DeepTSQP}, TRCF \cite{TRCF}, GMCL \cite{GMCL}, PLMF \cite{PLMF},  
                RNCF \cite{RNCF}, 
                STGCN \cite{stgcn}, 
                STGFT \cite{stgft},
                TaTruSR \cite{TaTruSR},
                \textbf{HCTN}},
                xtick=data,
                xticklabel style={rotate=75, anchor=east, font=\tiny},
                ymin=0,
                bar width=3pt,
                ymajorgrids=true,
                grid style=dashed  
            ]
            \addplot[fill=blue, draw=none] coordinates{
                (TUIPCC \cite{TUIPCC}, 7920)
                (BNLFT \cite{BNLFT}, 5900)
                (WSPred \cite{WSPRED}, 4600)
                (CTF \cite{CTF}, 4500)
                (NNCP \cite{NNCP}, 4400)
                (SCATSF \cite{SCATSF}, 4186.27)
                (TPMCF \cite{tpmcf}, 3934)
                (GFEN \cite{GFEN}, 3625.3861)
                (DeepTSQP \cite{DeepTSQP}, 3455.42)
                (TRCF \cite{TRCF}, 2736.095)
                (GMCL \cite{GMCL}, 1560.66)
                (PLMF \cite{PLMF}, 1532)
                (RNCF \cite{RNCF}, 1325.26)
                (STGCN \cite{stgcn}, 620.9300)
                (STGFT \cite{stgft}, 310.2)
                (TaTruSR \cite{TaTruSR}, 161.90)
                 (\textbf{HCTN}, 1460)     
                };
            \end{axis}
        \end{tikzpicture}
        \captionof{figure}{Training time comparison}
        \label{fig:train_time}
    \vspace{0.5cm}
        \begin{tikzpicture}
            \begin{axis}[
                width=\textwidth,
                height=0.6\textwidth,
                ylabel={Time (second)},
                ybar=5pt,
                symbolic x coords={TUIPCC \cite{TUIPCC}, TRCF \cite{TRCF}, SCATSF \cite{SCATSF}, TPMCF \cite{tpmcf}, TaTruSR \cite{TaTruSR}, RNCF \cite{RNCF}, DeepTSQP \cite{DeepTSQP}, 
                GMCL \cite{GMCL},   
                PLMF \cite{PLMF},  
                STGFT \cite{stgft},
                GFEN \cite{GFEN},  
                WSPred \cite{WSPRED}, NNCP \cite{NNCP}, 
                BNLFT \cite{BNLFT}, 
                CTF \cite{CTF}, 
                STGCN \cite{stgcn}, \textbf{HCTN}},
                xtick=data,
                xticklabel style={rotate=75, anchor=east, font=\tiny},
                ymode=log,
                bar width=3pt,
                ymajorgrids=true,
                grid style=dashed
            ]
            \addplot[fill=blue, draw=none] coordinates{
                (TUIPCC \cite{TUIPCC}, 0.002893)
                (TRCF \cite{TRCF}, 0.00058)
                (SCATSF \cite{SCATSF}, 0.00055774013)
                (TPMCF \cite{tpmcf}, 0.00041)
                (TaTruSR \cite{TaTruSR}, 0.00038651)
                (RNCF \cite{RNCF}, 0.0002401)
                (DeepTSQP \cite{DeepTSQP}, 0.0001774)
                (GMCL \cite{GMCL}, 0.00015)
                (PLMF \cite{PLMF}, 0.00015)
                (STGFT \cite{stgft}, 0.00005281602)
                (GFEN \cite{GFEN}, 0.000003388)
                (WSPred \cite{WSPRED}, 0.0000027990)
                (NNCP \cite{NNCP}, 0.0000027990)
                (BNLFT \cite{BNLFT}, 0.0000027990)
                (CTF \cite{CTF}, 7.94E-07)
                (STGCN \cite{stgcn}, 0.00000206109)
                (\textbf{HCTN}, 0.000002)};
            \end{axis}
        \end{tikzpicture}
        \captionof{figure}{Inference time comparison}
        \label{fig:test_time}
    \end{minipage}
\end{table*}

\noindent
{\textbf{Datasets}}: 
We evaluated our method on two publicly available QoS benchmark datasets from WSDREAM-2~\cite{WSDREAM}, comprising Response Time (RT) and Throughput (TP), denoted by D1 and D2, respectively. Table \ref{tab:dataset1} summarizes the dataset details.

\noindent
{\textbf{Train-Test-Validation Partition}}: 
{For both datasets, D1 (RT) and D2 (TP), we evaluate the model under five training densities $\psi \in \{5,10,15,20,50\}$. These are indexed as D1.$x$ and D2.$x$, where $x \in \{1,2,\dots,5\}$, following the mapping in Table~II; for example, D1.1 corresponds to D1 with $\psi = 5\%$. Since the datasets are temporal, we retain the first $\tau$ time-steps exactly as they appear in the original QoS tensor to serve as the historical input sequence. For the final $(\tau+1)^{\text{th}}$ time-step, we randomly sample $\psi\%$ of the valid QoS entries for training and use the remaining $(100-\psi)\%$ entries for testing; additionally, $20\%$ of the sampled training entries are reserved for validation. This sampling procedure is repeated five times for each $\psi$, and the reported results correspond to the average across these five independent train–test–validation splits, ensuring robustness against randomness and varying sparsity levels.
}

\noindent
{\textbf{Evaluation Metrics}}: 
To quantify prediction accuracy, HCTN is evaluated using Mean Absolute Error (MAE) and Root Mean Square Error (RMSE) \cite{CTF}. MAE reflects overall accuracy, while RMSE penalizes larger deviations more heavily. 


Appendix B provides dataset analysis, Appendix C lists hyperparameter settings, and Appendix D summarizes the comparison methods.


\noindent
{\textbf{Experimental Analysis}}: 
We now present an analysis of the performance of HCTN, starting with a comparison between HCTN and previous temporal QoS prediction methods.

\subsubsection{Performance Comparison of HCTN with Past Methods} 
\noindent
Table \ref{tab:soa_table} displays the performance of HCTN on the D1 and D2 datasets across five different training densities. HCTN significantly outperformed major previous methods in terms of both MAE and RMSE. The overall performance improvement relative to the \underline{second-best} method, denoted as $\mathcal{I}~(\%)$, is detailed in Table \ref{tab:soa_table}. 
Additionally, the results show HCTN performance improved with increased training density.

\subsubsection{Performance Comparison After Outlier Removal} 
We compared HCTN with all sixteen previous methods after removing 10\% of outliers (i.e., $\lambda{=}10$) from the D1 and D2 datasets at three different training densities, as detailed in Table \ref{tab:outlier_soa_table}. HCTN consistently achieved the lowest MAE and RMSE, demonstrating its superior predictive accuracy and robustness relative to other methods. This performance improvement is especially evident under challenging conditions, such as low training density, where the presence of outliers can severely distort the learning process, and removing even a small proportion of them leads to significant gains in prediction accuracy.

\subsubsection{Training Time of HCTN} HCTN was compared with all reported SOTA methods in terms of training time, as shown in Fig. \ref{fig:train_time}. Compared to TPMCF \cite{tpmcf}, CTF \cite{CTF}, and BNLFT \cite{BNLFT}, which previously achieved the best prediction accuracy across various cases, HCTN is faster by $1.87 \times$, $3.04 \times$, and $4.01 \times$, respectively. 

This demonstrates that HCTN can be retrained efficiently with minimal overhead as training density increases, resulting in enhanced prediction performance.

\subsubsection{Inference Time of HCTN} 
Real-time systems, especially those involving safety-critical applications, require swift QoS predictions. Fig. \ref{fig:test_time} shows that HCTN outperformed all the reported SOTA methods in terms of inference time. The average prediction time of HCTN was approximately $2 \times 10^{-6}$ seconds, a negligible duration compared to the minimum response time of services (0.001 second), as indicated in Table \ref{tab:dataset1}. This demonstrates that our method is well-suited for integration into real-time applications. 

Furthermore, while RNCF \cite{RNCF} exhibits a better training time compared to HCTN by $1.10 \times$, it falls short in inference time, which is in the range of $2.4{\times}10^{-4}$ (Fig.s \ref{fig:train_time}-\ref{fig:test_time}). TaTruSR \cite{TaTruSR}, being a memory-based method that employs only user similarity, also shows lower training time than HCTN. However, its inference time remains relatively high, in the order of $3.8{\times}10^{-4}$. In contrast, HCTN significantly outperformed both methods in terms of inference efficiency.
{Moreover, Appendix E in the supplementary file presents the asymptotic computational complexity of HCTN along with its trainable parameter count and FLOPs, offering a realistic estimate of its real-time efficiency.}

\begin{table}[!t]
    \centering
     \scriptsize 
    \caption{Comparative study of HCTN with $\lambda{=}10$}
    \begin{adjustbox}{width=0.85\linewidth}
        \begin{tabular}{l | c c c | c c c} \hline 
        \multirow{2}{*}{Methods} & \multicolumn{3}{c|}{MAE} & \multicolumn{3}{c}{RMSE} \\ \cline{2-7}
        & D1.2 & D1.4 & D1.5 & D1.2 & D1.4 & D1.5 \\\hline 
       
        WSPred \cite{WSPRED} & 2.1743 & 1.8934 & 1.8279 & 3.8617 & 3.6859 & 3.5229 \\ 
        TUIPCC \cite{TUIPCC} & 1.8622 & 1.6273 & 1.0306 & 4.4744 & 4.1678 & 2.9312 \\ 
        GFEN \cite{GFEN} & 1.7787 & 1.7775 & 1.5139 & 5.0724 & 4.9139 & 4.6031 \\
        SCATSF \cite{SCATSF} & 1.6076 & 1.5661 & 1.3333 & 4.8608 & 4.5484 & 3.6764\\ 
        DeepTSQP \cite{DeepTSQP} & 1.1732 & 1.0382 & 0.7216 & 3.5343 & 2.8141 & 1.9867 \\

        PLMF \cite{PLMF} & 1.1060 & 1.0316 & 0.9679 & 2.5411 & 2.4606 & 2.4003   \\ 
        BNLFT \cite{BNLFT} & 1.0828 & 1.0575  & 1.0475  & 2.6181 & 2.5809 & 2.5659  \\ 
        NNCP \cite{NNCP} & 1.0796 & 1.0536 & 1.0521 & 2.6401 & 2.5797 & 2.5805   \\  
        RNCF \cite{RNCF} & 1.0530 & 0.9482 & 0.8874 & 1.9867 & 2.4876 & 2.3733 \\ 
        TRCF \cite{TRCF} & 1.0354 & 0.7975 & 0.6995 & 3.6268 & 2.8328 & 2.1605 \\
        
        CTF \cite{CTF} & 0.9215 & 0.8981 & 0.8879 & 2.5865 & 2.5579 & 2.5541  \\ 

         STGCN \cite{stgcn} &  0.8663 &  0.8209 &  0.8191 &  2.8327 &  2.7455 &  2.7256 \\

        GMCL \cite{GMCL} & 0.8517 & 0.7860 & 0.7183  & 2.5405 & 2.4924 & 2.4723 \\
        
        TaTruSR \cite{TaTruSR} &  0.7742 & 0.6991 & 0.6316 & 2.4738 & 2.2084 & 2.0554 \\ 

         STGFT \cite{stgft} & 
         0.5728 &
         0.4298 &
         0.3266 &
         2.3098 &
         1.9686 &
         1.8577 \\

        TPMCF \cite{tpmcf} & \underline{0.4973} & \underline{0.3864} & \underline{0.1641} & \underline{2.2709} & \underline{1.8885} & \underline{1.3623} \\ \hline
        
        \textbf{HCTN} & \textbf{0.4713} & \textbf{0.3549} & \textbf{0.1569} & \textbf{1.9017} & \textbf{1.6893} & \textbf{1.1082} \\ \hline   
        $\mathcal{I} $ (\%) & 5.23\% & 8.15\% & 4.39\% & 16.26\% & 10.55\% & 18.65\%    \\  \hline \hline

        &  D2.2 & D2.4 & D2.5 & D2.2 & D2.4 & D2.5 \\\hline

        WSPred \cite{WSPRED} & 6.8200 & 6.2566 & 5.8448 & 34.7757 & 33.3952 & 33.0842 \\ 
        TUIPCC \cite{TUIPCC} & 6.3668 & 5.5349 & 4.0549 & 13.6695 & 11.3400 & 9.0899 \\ 
        GFEN \cite{GFEN} & 3.6144 & 3.4130 & 2.3500 & 5.6124 & 5.4232 & 4.4911 \\
        TaTruSR \cite{TaTruSR} & 3.0775 & 2.6818 & 2.2956 & 13.1061 & 10.3942 & 8.2938 \\

        DeepTSQP \cite{DeepTSQP} & 2.4883 & 4.1765 & {5.3630} & 5.7691 & {7.7521} & {9.4549} \\ 
        PLMF \cite{PLMF} & 2.4712 & 2.2602  & 2.3924 & 3.6705 & 3.8363 & 3.8347 \\ 
        SCATSF \cite{SCATSF} & 2.4281 & 2.1165 & 1.8738 & 4.9687 & 4.0603 & 3.6505 \\ 
        RNCF \cite{RNCF} & 1.8907 & 1.4847 & {1.5720} & 5.2260 & {4.8451} & 4.8397  \\
        
         STGCN \cite{stgcn} &  1.8283 &  1.4947 &  1.1833 &  4.4639 &  3.5102 &  2.8747 \\
        
        NNCP \cite{NNCP} & 1.5079 & 1.4342 & 1.3926 & 4.9207 & 4.7019 & 4.5026   \\ 
        TRCF \cite{TRCF} & 1.4895 & 1.0227 & 0.8414 & 4.9592 & 4.6396 & 3.9317 \\ 
        BNLFT \cite{BNLFT} & 1.4241 & 1.3935  & 1.3319 & 4.6031 & 4.4685 & 4.4319   \\

         STGFT \cite{stgft} & 
         1.4126 &
         1.2423 &
         1.1034 &
         3.1192 &
         2.8443 &
         2.6180 \\
        
        CTF \cite{CTF} & 1.3567 & 1.1945 & 1.0193 & 3.0436 & 2.9225 & 2.7989  \\ 
        GMCL \cite{GMCL} & 1.1780 & 1.1545 & 1.1222 & 3.7307 & 3.7236 & 3.5349 \\
        TPMCF \cite{tpmcf} & \underline{1.0101} & \underline{0.7881} & \underline{0.2210} & \underline{2.9564} & \underline{2.9092} & \underline{1.4372} \\ \hline
        \textbf{HCTN} & \textbf{0.9286} & \textbf{0.7204} & \textbf{0.2187}  & \textbf{2.9294} & \textbf{2.8475} & \textbf{1.1385} \\ \hline
        $\mathcal{I} $ (\%) & 8.07\% & 8.59\% & 1.04\% & 0.91\% & 2.12\% & 20.78\%   \\ \hline
        \end{tabular}
    \end{adjustbox}
    \label{tab:outlier_soa_table}
\end{table}
\begin{figure}[!b]
    \centering
    \begin{subfigure}{0.21\textwidth}\tiny
        \centering
          \begin{tikzpicture}
            \begin{axis}[
                width=\textwidth,
                height=\textwidth,
                view={45}{45},  
                xlabel={Training Densities},
                ylabel={Outliers (\%)},
                zlabel={MAE},
                xtick={0, 1, 2, 3},
                xticklabels={5\%, 10\%, 15\%, 20\%},
                ytick={0, 2, 4, 6, 8, 10},
                zmin=0.3, zmax=0.9,  
                ztick={0.3, 0.5, 0.7, 0.9},  
                zticklabels={0.3, 0.5, 0.7, 0.9},  
                colormap/jet,
                mesh/rows=6,  
                ymajorgrids=true,
                xmajorgrids=true,
                grid style=dashed,
                enlarge y limits={0.1},  
                colorbar,  
                colorbar style={
                    width=0.15cm,
                    ztick={0.3, 0.5, 0.7, 0.9},  
                }
            ]
            \addplot3[
                surf,  
                shader=interp,  
                ]
            coordinates {
                (0, 0, 0.8453)  (1, 0, 0.6765)  (2, 0, 0.6133)  (3, 0, 0.5256)
                (0, 2, 0.7574)  (1, 2, 0.6093)  (2, 2, 0.5495)  (3, 2, 0.4730)
                (0, 4, 0.7075)  (1, 4, 0.5670)  (2, 4, 0.5100)  (3, 4, 0.4373)
                (0, 6, 0.6625)  (1, 6, 0.5298)  (2, 6, 0.4746)  (3, 6, 0.4054)
                (0, 8, 0.6299)  (1, 8, 0.5008)  (2, 8, 0.4474)  (3, 8, 0.3811)
                (0, 10, 0.6081) (1, 10, 0.4713) (2, 10, 0.4282) (3, 10, 0.3549)  
            };
            \end{axis}
        \end{tikzpicture}
        \caption{\scriptsize{D1 Dataset}}
        \label{fig:outlier_rt_mae}
    \end{subfigure}
    ~~
    \begin{subfigure}{0.21\textwidth}\tiny
        \centering
          \begin{tikzpicture}
            \begin{axis}[
                width=\textwidth,
                height=\textwidth,
                view={45}{45},  
                xlabel={Training Densities},
                ylabel={Outliers (\%)},
                zlabel={MAE},
                xtick={0, 1, 2, 3},
                xticklabels={5\%, 10\%, 15\%, 20\%},
                ytick={0, 2, 4, 6, 8, 10},
                zmin=0.7, zmax=4.0,  
                ztick={0.7, 1.5, 2.7, 4.0},  
                zticklabels={0.7, 1.5, 2.7, 4.0},  
                colormap/jet,
                mesh/rows=6,  
                ymajorgrids=true,
                xmajorgrids=true,
                grid style=dashed,
                enlarge y limits={0.1},  
                colorbar,  
                colorbar style={
                    width=0.15cm,
                    ztick={0.7, 1.5, 2.7, 4.0},  
                }
            ]
             \addplot3[
                surf,  
                shader=interp,  
                ]
            coordinates {
                (0, 0, 3.9693)  (1, 0, 3.5486)  (2, 0, 3.3071)  (3, 0, 2.8671)
                (0, 2, 3.0269)  (1, 2, 2.2463)  (2, 2, 2.0754)  (3, 2, 1.6767)
                (0, 4, 2.2739)  (1, 4, 1.6144)  (2, 4, 1.5429)  (3, 4, 1.2411)
                (0, 6, 1.8131)  (1, 6, 1.2820)  (2, 6, 1.2066)  (3, 6, 0.9861)
                (0, 8, 1.5068)  (1, 8, 1.0720)  (2, 8, 1.0159)  (3, 8, 0.8243)
                (0, 10, 1.3166) (1, 10, 0.9286) (2, 10, 0.8207) (3, 10, 0.7204)
            };
            \end{axis}
        \end{tikzpicture}
         \caption{\scriptsize{D2 Dataset}}
        \label{fig:outlier_tp_mae}
    \end{subfigure}
    \caption{Impact of outliers}
    \label{fig:outlier_rt_tp}
\end{figure}

\subsubsection{Impact of Outliers} Fig.s \ref{fig:outlier_rt_tp}(a)-(b) depict the performance of HCTN in terms of MAE across varying levels of outlier removal ($\lambda=0$ to $\lambda=10$, in 2\% intervals) on the test datasets corresponding to D1 and D2, with varying training densities. 
Key observations indicate that as $\lambda$ increased, more outliers were detected and removed from the test datasets, leading to improved performance of HCTN.
However, the rate of improvement gradually decreased across all cases. 
The initial 2\% outlier removal yielded substantial performance gains, with average MAE improvements of 10.53\% for the D1 dataset and 34.80\% for the D2 dataset. 
In contrast, the performance gains became less pronounced in the final 2\% of outlier removal (from 8\% to 10\%), with only 4.65\% and 14.45\% average improvements in MAE for D1 and D2, respectively.

Analysis of the impact of outliers on HCTN performance showed that removing outliers significantly improved predictions, with the largest gains observed up to 6\% removal. Beyond this point, improvements taper off, indicating diminishing returns as the data becomes cleaner.


\subsubsection{Impact of Greysheep}
HCTN leverages LPAM to extract local features for individual users and services, combining these with the global feature representations from HCFM to effectively manage greysheep instances. Table \ref{tab:gs_outlier} shows that HCTN benefits from including local features for greysheep instances as opposed to excluding them (refer to HCTN - GMM in Table \ref{tab:gs_outlier}). 
On average, HCTN achieved a 2.93\% and 3.82\% improvement over HCTN-GMM on the D1 and D2 datasets, respectively.
Furthermore, applying local features to all users and services (refer to HCTN - GDM in Table \ref{tab:gs_outlier}) degraded performance. HCTN showed an average improvement of 1.59\% and 1.28\% over HCTN-GDM on D1 and D2, respectively. 
Although these improvements are modest, this is likely due to the small proportion of greysheep users (9.15\%) and services (11.51\%) identified, as only a limited fraction benefits from the GMM module. Nevertheless, addressing greysheep instances is critical for achieving more robust predictions in challenging cases.
This analysis highlights the importance of the GMM in selectively integrating local features to enhance accuracy.

Fig.s \ref{fig:gs_rt_tp}(a)-(b) illustrate the impact of the number of greysheep instances detected by the GDM and handled by the LPEM, which enhances their feature representation from HCFM by incorporating local patterns of users and services.
As the greysheep detection thresholds $c_1$ and $c_2$ decreased (here, $c_1 = c_2 = c$), the number of identified greysheep instances increased, initially resulting in improved HCTN performance. For example, when $c$ decreased from 3 to 1, HCTN achieved an average MAE improvement of 5.56\% on D1 and 2.29\% on D2, indicating that selectively enhancing more atypical users/services boosts prediction accuracy.
However, further lowering the threshold beyond $c = 1$ (e.g., $c = 0.2, 0.1$) led to performance degradation. This was due to an excessive number of users and services being labeled as greysheep, which overwhelmed the model with local patterns and diluted the benefits of collaborative filtering. Thus, while greysheep-aware enhancement improves accuracy up to a point, over-identification of greysheep instances can harm performance. These findings highlight the importance of appropriately setting the greysheep detection parameters to balance global and local pattern modeling for robust prediction. 

\begin{table}[!b]
    \centering
     \scriptsize
    \caption{Performance (MAE) of HCTN with selectively using LPAM with $c_1 = c_2 = 1$}
    \begin{adjustbox}{width=\linewidth}
    \begin{tabular}{c | c c c | c c c}
    \hline
      \multirow{2}{*}{Datasets}  & \multicolumn{3}{c|}{$\lambda=0$} & \multicolumn{3}{c}{$\lambda=10$} \\ \cline{2-7}
        & HCTN $-$ GMM &  HCTN $-$ GDM & \textbf{HCTN} &  HCTN $-$ GMM &  HCTN $-$ GDM & \textbf{HCTN} \\ \hline 
      D1.2 & 0.7036 & 0.6889 &  \textbf{0.6765} & 0.5055 & 0.4861 &   \textbf{0.4713}  \\ 
      D1.4 & 0.5364 & 0.5330 &  \textbf{0.5256} & 0.3669 & 0.3617 &  \textbf{0.3549}  \\ 
      D2.2 & 3.6713 & 3.5765 &  \textbf{3.5486} & 0.9550 & 0.9522 &  \textbf{0.9286}   \\
      D2.4 & 2.9956  & 2.9190 & \textbf{2.8671} & 0.7461 & 0.7451 &  \textbf{0.7204}  \\ \hline
      \multicolumn{7}{r}{HCTN $-$ GMM: HCTN without Greysheep Mitigation Module;}\\
      \multicolumn{7}{r}{HCTN $-$ GDM: HCTN without Greysheep Detection Module, fully utilizing LPAM}\\
    \end{tabular}
    \end{adjustbox}
    \label{tab:gs_outlier}
\end{table}

\begin{figure}[!t]
    \centering
    \begin{subfigure}{0.21\textwidth} 
        \centering 
         \scriptsize
        \begin{tikzpicture}
            \begin{axis}[
                width=1.25\textwidth,
                height=0.85\textwidth,
                ylabel={MAE},
                ybar=1pt,
                symbolic x coords={\shortstack{D1.2\\($\lambda=0$)}, \shortstack{D1.2\\($\lambda=10$)}, \shortstack{D1.4\\($\lambda=0$)}, \shortstack{D1.4\\($\lambda=10$)}},
                xticklabel style={rotate=45, anchor=east},
                xtick=data,
                legend style={at={(0.50,0.97)}, anchor=north, legend columns=3, /tikz/every even column/.append style={column sep=0.01cm},
                draw=none},
                legend image post style={scale=0.4},
                ymin=0.0,
                ymax=1.5,
                bar width=1.5pt,
                ymajorgrids=true,
                xmajorgrids=true,
                grid style=dashed
            ]

            \addplot [fill=blue, draw=none, draw opacity=0] coordinates {(\shortstack{D1.2\\($\lambda=0$)},0.6717) (\shortstack{D1.2\\($\lambda=10$)},0.4713) (\shortstack{D1.4\\($\lambda=0$)},0.5406) (\shortstack{D1.4\\($\lambda=10$)},0.3642)};

            \addplot [fill=red, draw=none, draw opacity=0] coordinates {(\shortstack{D1.2\\($\lambda=0$)},0.66691) (\shortstack{D1.2\\($\lambda=10$)},0.4699) (\shortstack{D1.4\\($\lambda=0$)},0.5360) (\shortstack{D1.4\\($\lambda=10$)},0.3580)};

            \addplot [fill=gray, draw=none, draw opacity=0] coordinates {(\shortstack{D1.2\\($\lambda=0$)},0.6558) (\shortstack{D1.2\\($\lambda=10$)},0.4614) (\shortstack{D1.4\\($\lambda=0$)},0.5252) (\shortstack{D1.4\\($\lambda=10$)},0.3538)};

            \addplot [fill=green, draw=none, draw opacity=0] coordinates {(\shortstack{D1.2\\($\lambda=0$)},0.6765) (\shortstack{D1.2\\($\lambda=10$)},0.4713) (\shortstack{D1.4\\($\lambda=0$)},0.5256) (\shortstack{D1.4\\($\lambda=10$)},0.3549)};
           
            \addplot [fill=cyan, draw=none, draw opacity=0] coordinates {(\shortstack{D1.2\\($\lambda=0$)},0.7023) (\shortstack{D1.2\\($\lambda=10$)},0.4938) (\shortstack{D1.4\\($\lambda=0$)},0.5518) (\shortstack{D1.4\\($\lambda=10$)},0.3785)};
            
            \addplot [fill=magenta, draw=none, draw opacity=0] coordinates {(\shortstack{D1.2\\($\lambda=0$)},0.7082) (\shortstack{D1.2\\($\lambda=10$)},0.497) (\shortstack{D1.4\\($\lambda=0$)},0.563) (\shortstack{D1.4\\($\lambda=10$)},0.3862)};
            
            \legend{\scalebox{0.75}{$c=0.1$}, \scalebox{0.75}{$c=0.2$}, \scalebox{0.75}{$c=0.5$}, \scalebox{0.75}{$c=1$}, \scalebox{0.75}{$c=2$}, \scalebox{0.75}{$c=3$}}
            \end{axis}
        \end{tikzpicture}
        \caption{\scriptsize{D1 Dataset}}
         \label{fig:gs_rt}
    \end{subfigure}
    ~~
   \begin{subfigure}{0.21\textwidth} 
        \centering
        \scriptsize
        \begin{tikzpicture}
            \begin{axis}[
                width=1.25\textwidth,
                height=0.85\textwidth,
                ylabel={MAE},
                ybar=1pt,
                symbolic x coords={\shortstack{D2.2\\($\lambda=0$)}, \shortstack{D2.2\\($\lambda=10$)}, \shortstack{D2.4\\($\lambda=0$)}, \shortstack{D2.4\\($\lambda=10$)}},
                xticklabel style={rotate=45, anchor=east},
                xtick=data,
                legend style={at={(0.50,0.97)}, anchor=north,legend columns=3, /tikz/every even column/.append style={column sep=0.01cm},
                draw=none},
                legend image post style={scale=0.4},
                ymin=0.0,
                ymax=8,
                bar width=1.5pt,
                ymajorgrids=true,
                xmajorgrids=true,
                grid style=dashed
            ]
            \addplot [fill=blue, draw=none, draw opacity=0] coordinates {(\shortstack{D2.2\\($\lambda=0$)},3.4745) (\shortstack{D2.2\\($\lambda=10$)},0.9003) (\shortstack{D2.4\\($\lambda=0$)},3.0901) (\shortstack{D2.4\\($\lambda=10$)},0.7830)};

            \addplot [fill=red, draw=none, draw opacity=0] coordinates {(\shortstack{D2.2\\($\lambda=0$)},3.4327) (\shortstack{D2.2\\($\lambda=10$)},0.8570) (\shortstack{D2.4\\($\lambda=0$)},3.0579) (\shortstack{D2.4\\($\lambda=10$)},0.7495)};

            \addplot [fill=gray, draw=none, draw opacity=0] coordinates {(\shortstack{D2.2\\($\lambda=0$)},3.3982) (\shortstack{D2.2\\($\lambda=10$)},0.8499) (\shortstack{D2.4\\($\lambda=0$)},2.8881) (\shortstack{D2.4\\($\lambda=10$)},0.7094)};
            
            \addplot [fill=green, draw=none, draw opacity=0] coordinates {(\shortstack{D2.2\\($\lambda=0$)},3.5486) (\shortstack{D2.2\\($\lambda=10$)},0.9286) (\shortstack{D2.4\\($\lambda=0$)},2.8671) (\shortstack{D2.4\\($\lambda=10$)},0.7204)};
            
            \addplot [fill=cyan, draw=none, draw opacity=0] coordinates {(\shortstack{D2.2\\($\lambda=0$)},3.6134) (\shortstack{D2.2\\($\lambda=10$)},0.9558) (\shortstack{D2.4\\($\lambda=0$)},2.8926) (\shortstack{D2.4\\($\lambda=10$)},0.7374)};
            
            \addplot [fill=magenta, draw=none, draw opacity=0] coordinates {(\shortstack{D2.2\\($\lambda=0$)},3.6161) (\shortstack{D2.2\\($\lambda=10$)},0.9625) (\shortstack{D2.4\\($\lambda=0$)},2.9472) (\shortstack{D2.4\\($\lambda=10$)},0.7425)};

            \legend{\scalebox{0.75}{$c=0.1$}, \scalebox{0.75}{$c=0.2$}, \scalebox{0.75}{$c=0.5$}, \scalebox{0.75}{$c=1$}, \scalebox{0.75}{$c=2$}, \scalebox{0.75}{$c=3$}}
            \end{axis}
        \end{tikzpicture}
        \caption{\scriptsize{D2 Dataset}}
        \label{fig:gs_tp}
    \end{subfigure}
     \caption{Impact of greysheep}
    \label{fig:gs_rt_tp}
\end{figure}

\subsubsection{Module Ablation Study for HCTN} 
This analysis justifies the necessity of each module within HCTN by highlighting their contributions to overall performance, measured by MAE, across various $\lambda$ values on the D1 and D2 datasets. 
As shown in Table \ref{tab:module_study}, including HCFM (i.e., HCTN $-$ TGEM) rather than excluding it (i.e., HCTN $-$ HCFM $-$ TGEM) resulted in an average improvement of 2.51\% on D1 and 6.11\% on D2. 
Similarly, adding TGEM (i.e., HCTN $-$ HCFM) rather than ablating it (i.e., HCTN $-$ HCFM $-$ TGEM) led to a 1.23\% and 2.22\% improvement on D1 and D2, respectively. 
Combining both TGEM and HCFM (i.e., HTCN) yielded even better results, with an average improvement of 1.59\% on D1 and 5.44\% on D2 over the HCTN $-$ TGEM setup. 
Furthermore, HCTN showed a notable improvement of 3.73\% on D1 and 8.43\% on D2 over HCTN $-$ HCFM, 
 highlighting the importance of the HCFM and TGEM modules in optimizing performance.

\begin{table}[!t]
    \centering \scriptsize
     \caption{Module ablation study for HCTN (MAE)}
     \begin{adjustbox}{width=0.45\textwidth}
        \begin{tabular}{c | l | c c c c c c}
            \hline 
           \multirow{2}{*}{Datasets} & \multirow{2}{*}{Modules}  & \multicolumn{6}{c}{$\lambda$}  \\ \cline{3-8}
           & & $0$ & $2$ & $4$ & $6$ & $8$ & $10$ \\ \hline 
           \multirow{4}{*}{D1.2} 
           & HCTN $-$ HCFM $-$ TGEM &      0.7075	& 0.6335	& 0.5896 	& 0.5511	& 0.5223	& 0.5027 \\ 
           & HCTN $-$ TGEM & 0.6995	& 0.623	&    0.5803	& 0.5424	& 0.5129	& 0.4925	\\ 
           & HCTN $-$ HCFM & 0.7058	& 0.6299 &  0.5882 & 0.5507	& 0.521	& 0.5008 \\ 
           & HCTN &       0.6765 & 0.6093 &     0.5670 & 0.5298 & 0.5008 & 0.4713  \\ \hline 
    
           \multirow{4}{*}{D1.4} 
           & HCTN $-$ HCFM $-$ TGEM &     0.5559 & 0.4924 & 0.4577 & 0.4267 & 0.4029 & 0.3866	\\ 
           & HCTN $-$ TGEM & 0.5343	& 0.4892	& 0.4536	& 0.4221	& 0.3981	& 0.3814	\\ 
           & HCTN $-$ HCFM & 0.5436	& 0.4798	& 0.4437	& 0.4114	& 0.3869	& 0.3698	\\ 
           & HCTN & 0.5256	& 0.473	& 0.4373	& 0.4054	& 0.3811	& 0.3549	\\ \hline 
    
           \multirow{4}{*}{D2.2} 
           & HCTN $-$ HCFM $-$ TGEM & 4.1140	& 2.3369	& 1.6610 & 1.3194	& 1.1099	& 0.9642	\\ 
           & HCTN $-$ TGEM & 3.8710	& 2.3314	&  1.6557	& 1.3169	& 1.1018	& 0.9550	\\ 
           & HCTN $-$ HCFM & 3.9846	& 2.3325	& 1.6372	& 1.3002	& 1.0904	& 0.9474	\\ 
           & HCTN & 3.5486 & 2.2463 & 1.6144 & 1.2820 & 1.0720 & 0.9286	\\ \hline 

           \multirow{4}{*}{D2.4} 
           & HCTN $-$ HCFM $-$ TGEM & 3.0876 & 1.7567 & 1.3180 & 1.0327	& 0.8655	& 0.7562	\\ 
           & HCTN $-$ TGEM &  2.8926	& 1.7051 & 1.2810 & 1.0289	& 0.8565	& 0.7476	\\
           & HCTN $-$ HCFM & 3.0475 & 1.7269 & 1.2865 & 1.0461	& 0.8514	& 0.7446	\\ 
           & HCTN & 2.8671 & 1.6767	& 1.2422	& 0.9876	& 0.8288	& 0.7204 \\ \hline 

        \end{tabular}
    \end{adjustbox}
    \label{tab:module_study}
\end{table}

\begin{figure}[!b] 
    \centering
    \begin{subfigure}{0.20\textwidth} 
        \centering 
        \scriptsize
        \begin{tikzpicture}
            \begin{axis}[
                width=1.25\textwidth,
                height=0.80 \textwidth,
                ylabel={MAE},
                ybar=1pt,
                symbolic x coords={\shortstack{D1.2\\($\lambda=0$)}, \shortstack{D1.2\\($\lambda=10$)}, \shortstack{D1.4\\($\lambda=0$)}, \shortstack{D1.4\\($\lambda=10$)}},
                xticklabel style={rotate=45, anchor=east},
                xtick=data,
                legend style={at={(0.5,0.97)}, anchor=north,legend columns=2, /tikz/every even column/.append style={column sep=-0.2cm}, 
                draw=none},
                legend image post style={scale=0.50},
                ymin=0,
                ymax=1.6,
                ytick={0, 0.4, 0.8, 1.2, 1.6},
                bar width=3pt,
                ymajorgrids=true,
                xmajorgrids=true,
                grid style=dashed
            ]
           \addplot [fill=blue, draw=none, draw opacity=0] coordinates {(\shortstack{D1.2\\($\lambda=0$)}, 0.7163) (\shortstack{D1.2\\($\lambda=10$)}, 0.5050)(\shortstack{D1.4\\($\lambda=0$)}, 0.5739) (\shortstack{D1.4\\($\lambda=10$)}, 0.3908)};

           \addplot [fill=red, draw=none, draw opacity=0] coordinates {(\shortstack{D1.2\\($\lambda=0$)},0.6849) (\shortstack{D1.2\\($\lambda=10$)},0.4858) (\shortstack{D1.4\\($\lambda=0$)},0.5327) (\shortstack{D1.4\\($\lambda=10$)},0.3606)};
           
           \addplot [fill=green, draw=none, draw opacity=0] coordinates {(\shortstack{D1.2\\($\lambda=0$)},0.6765) (\shortstack{D1.2\\($\lambda=10$)},0.4713)
           (\shortstack{D1.4\\($\lambda=0$)},0.5256) (\shortstack{D1.4\\($\lambda=10$)},0.3549)};
            
            \legend{\scalebox{0.60}{FCU}, \scalebox{0.60}{SUCU$+$SSCU}, \scalebox{0.60}{FCU$+$SUCU$+$SSCU}}
            \end{axis}
        \end{tikzpicture}
        \caption{\scriptsize{D1 Dataset}}
         \label{fig:gconv_rt}
    \end{subfigure}
    ~~
    \begin{subfigure}{0.20\textwidth} 
        \centering
        \scriptsize
        \begin{tikzpicture}
            \begin{axis}[
                width=1.25\textwidth,
                height=0.80 \textwidth,
                ylabel={MAE},
                ybar=1pt,
                symbolic x coords={\shortstack{D2.2\\($\lambda=0$)}, \shortstack{D2.2\\($\lambda=10$)}, \shortstack{D2.4\\($\lambda=0$)}, \shortstack{D2.4\\($\lambda=10$)}},
                xticklabel style={rotate=45, anchor=east},
                xtick=data,
                legend style={at={(0.5,0.97)}, anchor=north,legend columns=2, /tikz/every even column/.append style={column sep=-0.2cm},
                draw=none},
                legend image post style={scale=0.50},
                ymin=0.0,
                ymax=7.5,
                ytick={0, 2.5, 5.0, 7.5},
                bar width=3pt,
                ymajorgrids=true,
                xmajorgrids=true,
                grid style=dashed
            ]
           
            \addplot [fill=blue, draw=none, draw opacity=0] coordinates {(\shortstack{D2.2\\($\lambda=0$)},4.1295) (\shortstack{D2.2\\($\lambda=10$)},0.9730) (\shortstack{D2.4\\($\lambda=0$)},3.3481) (\shortstack{D2.4\\($\lambda=10$)},0.7348)};

            \addplot [fill=red, draw=none, draw opacity=0] coordinates {(\shortstack{D2.2\\($\lambda=0$)},3.8843) (\shortstack{D2.2\\($\lambda=10$)},0.9625) (\shortstack{D2.4\\($\lambda=0$)},3.0661) (\shortstack{D2.4\\($\lambda=10$)},0.7325)};

            \addplot [fill=green, draw=none, draw opacity=0] coordinates {(\shortstack{D2.2\\($\lambda=0$)},3.5486) (\shortstack{D2.2\\($\lambda=10$)},0.9286) (\shortstack{D2.4\\($\lambda=0$)},2.8671) (\shortstack{D2.4\\($\lambda=10$)},0.7204)};
            
             \legend{\scalebox{0.60}{FCU}, \scalebox{0.60}{SUCU$+$SSCU}, \scalebox{0.60}{FCU$+$SUCU$+$SSCU}}
            \end{axis}
        \end{tikzpicture}
         \caption{\scriptsize{D2 Dataset}}
        \label{fig:gconv_tp}
    \end{subfigure}
     \caption{Impact of hypergraph}
    \label{fig:hypergraph_component}
\end{figure}

\subsubsection{Feature Ablation Study: Impact of Higher-order Graphs Features}
This analysis highlights the significance of integrating higher-order collaborative features extracted from hypergraphs as defined in this paper. 
HCFM is designed to extract features from FIG, as well as from SUIG and SSIG. 
FIG captures features from heterogeneous nodes, while SUIG and SSIG focus on homogeneous user-user and service-service node interactions, respectively. Fig. \ref{fig:hypergraph_component} illustrates that combining features from FIG, SUIG, and SSIG yielded superior performance compared to using FIG alone. The combined features resulted in an average improvement of 6.99\% on the D1 dataset and 19.06\% on the D2 dataset over the features extracted from FIG alone. Additionally, the combined features outperformed those extracted from SUIG and SSIG by an average of 1.28\% on D1 and 17.88\% on D2, highlighting the
enhanced value of complex feature integration enabled by the introduction of QIHG.

\begin{figure}[!t] 
    \centering
    \begin{subfigure}{0.21\textwidth} 
        \centering 
        \scriptsize
        \begin{tikzpicture}
            \begin{axis}[
                width=1.26\textwidth,
                height=0.85 \textwidth,
                ylabel={MAE},
                ybar=1pt,
                symbolic x coords={\shortstack{D1.2\\($\lambda=0$)}, \shortstack{D1.2\\($\lambda=10$)}, \shortstack{D1.4\\($\lambda=0$)}, \shortstack{D1.4\\($\lambda=10$)}},
                xticklabel style={rotate=45, anchor=east},
                xtick=data,
                legend style={at={(0.58,0.97)}, anchor=north,legend columns=2, /tikz/every even column/.append style={column sep=0.01cm},
                draw=none},
                legend image post style={scale=0.50},
                ymin=0.3,
                ymax=0.9,
                ytick={0.3, 0.6, 0.9},
                bar width=3pt,
                ymajorgrids=true,
                xmajorgrids=true,
                grid style=dashed
            ]
            
            \addplot [fill=blue, draw=none, draw opacity=0] coordinates {(\shortstack{D1.2\\($\lambda=0$)},0.7249) (\shortstack{D1.2\\($\lambda=10$)},0.5031) (\shortstack{D1.4\\($\lambda=0$)},0.5572) (\shortstack{D1.4\\($\lambda=10$)},0.3899)};
           
           \addplot [fill=red, draw=none, draw opacity=0] coordinates {(\shortstack{D1.2\\($\lambda=0$)},0.7209) (\shortstack{D1.2\\($\lambda=10$)},0.5017) (\shortstack{D1.4\\($\lambda=0$)},0.5491) (\shortstack{D1.4\\($\lambda=10$)},0.3752)};
           
            \addplot [fill=gray, draw=none, draw opacity=0] coordinates {(\shortstack{D1.2\\($\lambda=0$)},0.7156) (\shortstack{D1.2\\($\lambda=10$)},0.5006) (\shortstack{D1.4\\($\lambda=0$)},0.5452) (\shortstack{D1.4\\($\lambda=10$)},0.3731)};
            
            \addplot [fill=green, draw=none, draw opacity=0] coordinates {(\shortstack{D1.2\\($\lambda=0$)},0.6765) (\shortstack{D1.2\\($\lambda=10$)},0.4713) (\shortstack{D1.4\\($\lambda=0$)},0.5256) (\shortstack{D1.4\\($\lambda=10$)},0.3549)};
            
             \legend{\scalebox{0.70}{E block}, \scalebox{0.70}{E$+$T block}, \scalebox{0.70}{E$+$F block}, \scalebox{0.70}{E$+$T$+$F block}}
            \end{axis}
        \end{tikzpicture}
         \caption{\scriptsize{D1 Dataset}}
         \label{fig:t_f_rt}
    \end{subfigure}
    ~~
    \begin{subfigure}{0.21\textwidth} 
        \centering
        \scriptsize
        \begin{tikzpicture}
            \begin{axis}[
                width=1.26\textwidth,
                height=0.85 \textwidth,
                ylabel={MAE},
                ybar=1pt,
                symbolic x coords={\shortstack{D2.2\\($\lambda=0$)}, \shortstack{D2.2\\($\lambda=10$)}, \shortstack{D2.4\\($\lambda=0$)}, \shortstack{D2.4\\($\lambda=10$)}},
                xticklabel style={rotate=45, anchor=east},
                xtick=data,
                legend style={at={(0.58,0.97)}, anchor=north,legend columns=2, /tikz/every even column/.append style={column sep=0.01cm},
                draw=none},
                legend image post style={scale=0.50},
                ymin=0.0,
                ymax=6,
                bar width=3pt,
                ymajorgrids=true,
                xmajorgrids=true,
                grid style=dashed
            ]
	
            \addplot [fill=blue, draw=none, draw opacity=0] coordinates {(\shortstack{D2.2\\($\lambda=0$)},3.8933) (\shortstack{D2.2\\($\lambda=10$)},0.9730) (\shortstack{D2.4\\($\lambda=0$)},3.0653) (\shortstack{D2.4\\($\lambda=10$)},0.7602)};
            
            \addplot [fill=red, draw=none, draw opacity=0] coordinates {(\shortstack{D2.2\\($\lambda=0$)},3.8086) (\shortstack{D2.2\\($\lambda=10$)},0.9550) (\shortstack{D2.4\\($\lambda=0$)},2.9800) (\shortstack{D2.4\\($\lambda=10$)},0.7338)};
            
            \addplot [fill=gray, draw=none, draw opacity=0] coordinates {(\shortstack{D2.2\\($\lambda=0$)},3.7645) (\shortstack{D2.2\\($\lambda=10$)},0.9392) (\shortstack{D2.4\\($\lambda=0$)},2.9350) (\shortstack{D2.4\\($\lambda=10$)},0.7247)};
            
            \addplot [fill=green, draw=none, draw opacity=0] coordinates {(\shortstack{D2.2\\($\lambda=0$)},3.5486) (\shortstack{D2.2\\($\lambda=10$)},0.9286) (\shortstack{D2.4\\($\lambda=0$)},2.8671) (\shortstack{D2.4\\($\lambda=10$)},0.7204)};
            
             \legend{\scalebox{0.70}{E block}, \scalebox{0.70}{E$+$T block}, \scalebox{0.70}{E$+$F block}, \scalebox{0.70}{E$+$T$+$F block}}
            \end{axis}
        \end{tikzpicture}
         \caption{\scriptsize{D2 Dataset}}
        \label{fig:t_f_tp}
    \end{subfigure}
     \caption{Impact of various blocks in TGEM}
    \label{fig:t_f_block}
\end{figure}

\subsubsection{Feature Ablation Study: Impact of Multi-granularity Temporal Features} 
This analysis demonstrates the value of integrating multi-granularity temporal features using TGEM, which includes the E block, T block, and F block. The T block specializes in capturing fine-grained temporal features, while the F block focuses on coarse-grained features. Fig. \ref{fig:t_f_block} illustrates that combining both fine-grained and coarse-grained features yielded better performance on the D1 and D2 datasets compared to using each feature type individually.
The combined features from the E $+$ T $+$ F blocks provided an average improvement of 6.58\% on the D1 dataset and 8.17\% on the D2 dataset over the features extracted by the E block alone. Similarly, E $+$ T $+$ F blocks resulted in a 5.52\% and 5.63\% improvement on the D1 and D2 datasets, respectively, compared to the E $+$ T block setup. Furthermore, E $+$ T $+$ F blocks outperformed the E $+$ F block combination by 4.53\% on D1 and 4.02\% on D2, underscoring the more significant impact of incorporating more comprehensive and diverse feature sets. 


\subsubsection{Statistical Analysis}
We now present the statistical analysis of the performance of HTCN. Table \ref{tab:ci} displays the analysis across all datasets, using three different confidence limits (CL): 90\%, 95\%, and 99\%. For example, a CL of 90\% indicates that the given samples of QoS values are 90\% likely to contain the population's true mean. The confidence interval analysis shows that as the CL increases, the range of the interval broadens, indicating greater certainty about capturing the true mean of QoS values, but with a wider range. This reflects a trade-off between precision and confidence in the performance estimates of HCTN.

\begin{table}[!t] 
    \centering
    \scriptsize
    \caption{Statistical analysis using confidence intervals}
    \begin{adjustbox}{width=0.85\linewidth}
    \begin{tabular}{c | c  c c c } \hline
         CL & D1.2 & D1.4 & D2.2 & D2.4 \\ \hline
          90\% & (0.6812, 0.6904) & (0.5214, 0.5296) & (3.5424, 3.5548) & (2.8166, 2.9176) \\ 
          
          95\% & (0.6804, 0.6913) &  (0.5206, 0.5304) & (3.5412, 3.5560) & (2.8069, 2.9273) \\ 
          
          99\% & (0.6786, 0.6930) &  (0.5191, 0.5320) & (3.5389, 3.5584) & (2.7880, 2.9462) \\  \hline

          Mean  & 0.6765 & 0.5256 & 3.5486	& 2.8671 \\ 
          Std. Dev. & 0.0219 & 0.0207 & 0.1891 & 0.1592 \\ \hline
    \end{tabular}
    \end{adjustbox}
    \label{tab:ci}
\end{table}

HCTN demonstrates superior performance compared to previous methods, with significant improvements in MAE and RMSE across various datasets and training densities. It also excels in handling outliers, provides efficient training and prediction times, and benefits from advanced feature integration, including higher-order graph features and multi-granularity temporal features. 
A more detailed experimental analysis, including the impact of cold-start and the influence of various hyperparameters on the performance of HCTN, is presented in Appendices F and G, respectively. {Additionally, Appendix H presents a study on the deployability of HCTN in resource-constrained environments.}
\section{Literature Review} \label{sec:lit_review}

\noindent
QoS prediction has been extensively studied for service composition and recommendation. Classical approaches are largely based on collaborative filtering (CF), categorized into: 
(a) \emph{Memory-based}, using user-service correlations \cite{UPCC,IPCC,WSRec}; 
(b) \emph{Model-based}, learning parameters via mathematical models \cite{MF,NMF}; and 
(c) \emph{Hybrid}, combining both for better accuracy \cite{trqp,OFFDQ}. 
However, many existing methods \cite{UPCC,IPCC,WSRec,NMF,trqp} assume static QoS, ignoring temporal variations. 
Despite advancements, challenges like data sparsity, credibility, and robust learning remain. 

\begin{table}[!t]
    \centering
    \caption{Literature survey on temporal QoS prediction}
    \begin{adjustbox}{width=\linewidth}
    \scriptsize
    \begin{tabular}{l | l | c c | c c | c c}
        \hline
         \multirow{2}{*}{Methods} & \multirow{2}{*}{Techniques} & \multicolumn{2}{c|}{Data Deficiency} & \multicolumn{2}{c|}{Data Credibility} & \multicolumn{2}{c}{Data Representation}    \\ \cline{3-8}
         & & RQ1 & RQ2 & RQ3 & RQ4
         & RQ5 & RQ6 \\ \hline
         \cite{TASR, TF-KMP} & ARIMA &  \xmark &  \xmark & \xmark & \xmark & \cmark & \xmark \\ \hline
        
        \cite{TRCF, TUIPCC, etacf} & \multirow{3}{*}{Temporal Average, Smoothing, Decay} & \xmark & \xmark & \xmark & \xmark & \cmark & \xmark \\   
         \cite{CASR-TSE} &  & \cmark & \cmark & \xmark & \xmark & \cmark & \xmark \\    
         \cite{TaTruSR} &  & \cmark & \cmark & \xmark & \pmark & \cmark & \xmark \\ \hline  

         \cite{AMF} & \multirow{2}{*}{Adaptive MF} & \cmark & \cmark & \pmark & \xmark & \cmark & \xmark \\  
         \cite{TMF} &  & \cmark & \cmark & \xmark & \xmark & \cmark & \xmark \\ \hline 
        
        \cite{BNLFT, WSPRED,NNCP}  & \multirow{3}{*}{Tensor Factorization} & \cmark & \cmark & \pmark & \xmark & \cmark  & \xmark \\  
        \cite{BNTucF, RNLFT, dnl, HDOP, INAL, RNL}  &  & \cmark & \cmark & \pmark & \xmark & \cmark  & \xmark \\  

         \cite{CTF} &  & \cmark & \cmark & \cmark & \xmark & \cmark  & \xmark \\ \hline 

         \cite{Mul-TSFL},  \cite{PLMF} &  \multirow{2}{*}{Long-Short Term Memory (LSTM)} & \cmark & \cmark & \xmark & \xmark & \cmark & \cmark \\ 
         \cite{QSPC} &  & \cmark &  \cmark & \pmark & \xmark &  \cmark & \cmark \\ \hline 
        
         \cite{DeepTSQP} & \multirow{4}{*}{Gated Recurrent Units (GRU)} &  \xmark & \xmark & \xmark & \xmark & \cmark & \cmark \\  
         \cite{GFEN} &  & \cmark & \cmark & \xmark & \xmark & \cmark & \cmark \\  
         \cite{RNCF} &  & \cmark & \cmark & \xmark & \xmark & \xmark & \cmark \\  
         \cite{SCATSF},  \cite{RTF} &  &  \cmark &  \cmark & \pmark & \xmark & \cmark & \cmark \\ \hline   
        
     
         \cite{clpred} & \multirow{2}{*}{Transformer} &  \cmark &  \cmark & \pmark & \xmark & \cmark  & \cmark \\ 
         \cite{MFInformer} &  & \cmark & \cmark & \xmark & \xmark & \cmark  & \cmark \\ \hline

         \cite{GMCL} & Graph Convolution + CL  &  \cmark & \xmark & \pmark & \xmark & \xmark & \cmark \\\hline

         \cite{dgncl} & Graph Attention + GRU  &  \cmark & \xmark & \pmark & \xmark & \xmark & \cmark \\\hline
         \cite{tpmcf} & Graph Convolution + Transformer & \cmark & \cmark & \cmark & \xmark & \cmark & \cmark \\\hline
       
        {HCTN} (ours) & Hypergraph Convoluted Transformer  &  \cmark &  \cmark & \cmark & \cmark & \cmark & \cmark \\ \hline
        
    \multicolumn{8}{r}{\cmark: Addressed, \pmark: Partially addressed, \xmark: Not addressed}   
    \end{tabular}
    \end{adjustbox}
    \label{tab:literature_survey}
\end{table}

\subsubsection{Architectural Solutions for Time-Aware QoS Prediction}
Temporal QoS prediction methods fall into two main categories: memory-based and model-based. Memory-based methods are simple and intuitive but computationally expensive on large datasets. In contrast, model-based methods, such as Tensor Factorization (TF), Adaptive MF, ARIMA, RNNs, and Transformers, offer better scalability and accuracy in capturing QoS dynamics over time. Early neighborhood methods used time-slice averaging. Tong et al. \cite{TUIPCC} improved this by filtering outdated data. Time-decay strategies were later applied in \cite{etacf, TaTruSR} to emphasize recent QoS. \cite{CASR-TSE} added spatial context, and \cite{TRCF} improved similarity metrics using RBS and PCC.

To address limitations of neighborhood-based methods, model-based approaches emerged. 
\cite{WSPRED} introduced low-rank TF with QoS regularization, extended by \cite{NNCP} via non-negativity and CP decomposition. 
\cite{CTF} proposed closed-form updates under KKT conditions, while \cite{RNLFT} tackled overfitting using multiple regularizations. Tucker decomposition was used in \cite{BNTucF}, and adjustable-depth networks in \cite{dnl}. Further enhancements include multi-dimensional modeling \cite{HDOP}, robust TF with ADMM \cite{RNL}, and density-aware TF with PSO \cite{INAL}. Adaptive MF methods like \cite{AMF, TMF} improved temporal modeling via online SGD and temporal smoothing. ARIMA-based models \cite{TASR, TF-KMP} captured linear trends but assume time-series stationarity, which may not hold in dynamic environments.

RNN-based deep learning models like LSTM and GRU were later applied. \cite{PLMF} combined LSTM with MF, and \cite{Mul-TSFL} used multivariate LSTM with spatial context. \cite{QSPC} incorporated contextual information, while \cite{RNCF, DeepTSQP, GFEN, SCATSF, RTF} extended GRU-based architectures for sequence modeling. While RNNs process sequentially, Transformers model long-range dependencies in parallel. \cite{clpred} used a contrastive learning framework with Transformers and augmented QoS sequences; \cite{MFInformer} combined MF with Informer for temporal prediction. Recently, GNNs have also been explored: \cite{GMCL} proposed tripartite contrastive graph learning; \cite{dgncl} applied GRUs on dynamic QoS graphs; and \cite{tpmcf} integrated graph convolutions for spatial and Transformers for temporal modeling.
We now examine these architectures in the context of key challenges through six research questions (\emph{RQs}) addressed in this paper.

\subsubsection{Data Deficiency Problem} 
The QoS data deficiency problem arises mainly from (a) sparse user-service interactions and (b) cold-start scenarios, where historical data is missing for new users or services. Similarity-based methods \cite{TASR, TRCF, TUIPCC, etacf, DeepTSQP} often yield low accuracy in such cases. CLpred \cite{clpred} addresses sparsity via data imputation but struggles with cold-start issues. Factorization approaches like MF \cite{MF, PLMF, tpmcf, AMF, TMF, GFEN, MFInformer} and TF \cite{WSPRED, CTF, RNL, BNLFT, BNTucF, RNLFT, dnl, HDOP, INAL, NNCP, RTF} integrate metadata (e.g., user/service location) \cite{Mul-TSFL, QSPC, SCATSF} to alleviate these issues. However, such metadata may be unavailable due to privacy concerns, and MF/TF models often miss higher-order features and credibility challenges, affecting prediction accuracy.

\subsubsection{Data Credibility Problem}
Credibility issues in QoS data arise mainly from outliers and greysheep. 
(a) \emph{Outliers:}  
These are extreme QoS values caused by service/network anomalies. Many methods \cite{TASR, TRCF, TUIPCC, etacf, TMF, DeepTSQP, MFInformer} ignore outliers, leading to poor accuracy. Solutions include  
(i) unsupervised detection (e.g., isolation forest \cite{CTF, tpmcf}), and  
(ii) robust loss functions like Cauchy \cite{CTF, RNL}, Huber \cite{RTF}, MAE \cite{QSPC}, or noise contrastive loss \cite{clpred}. Regularization techniques (L1, L2, etc.) also help mitigate outlier effects \cite{dgncl, BNLFT, AMF, WSPRED}. 
(b) \emph{Greysheep:}  
Users/services with atypical patterns often hinder prediction. While clustering \cite{CAP} and reputation-based \cite{RAP, TAP} methods attempt to handle them, these approaches may overlook their unique value and fail to generalize well.

\subsubsection{Data Representation Learning Problem}
Effective QoS prediction depends on data representations that capture relevant patterns. Two major types are commonly used: 
(a) \emph{Domain-specific features:} There are primarily two distinct types of domain-specific features. 
(i) QoS features—user-service similarities—are used in \cite{TASR, TRCF, TUIPCC, etacf, DeepTSQP, TaTruSR} but suffer from sparsity and cold-start issues. MF \cite{AMF, TMF} and TF \cite{CTF, BNLFT, WSPRED, BNTucF, HDOP, INAL, NNCP} alleviate sparsity but struggle to capture higher-order interactions.  
(ii) Contextual features, 
like metadata (e.g., IP addresses, locations) in \cite{Mul-TSFL, QSPC, SCATSF}, 
aid cold-start handling but raise privacy concerns. 
(b) \emph{Higher-order features:}
Advanced models like MLPs, GNNs, and RNNs (LSTM/ GRU) capture non-linear, higher-order dependencies. Works such as \cite{dgncl, tpmcf} leverage GNNs and transformers for spatial-temporal modeling, while \cite{Mul-TSFL, PLMF, QSPC, DeepTSQP, GFEN, SCATSF, RTF, clpred, MFInformer} combine these with domain-specific features to enhance prediction accuracy.

Overall, QoS prediction methods often struggle to capture higher-order interactions, handle non-linearity, and deal with cold-start and data sparsity. Domain-specific features may be limited by high dimensionality and privacy concerns, while advanced models can be computationally intensive and lack scalability. 
In contrast, our HCTN integrates domain features with deep learning via a hypergraph-convoluted transformer. By combining non-negative matrix decomposition with robust architectures, HCTN addresses data deficiency, cold-start, and credibility issues, capturing complex non-linear patterns and improving both accuracy and scalability. Table~\ref{tab:literature_survey} compares HCTN with SOTA methods based on the key QoS prediction challenges addressed.
\section{Conclusion} \label{sec:conclusion}

\noindent
This paper proposes HCTN, a Hypergraph Convoluted Transformer Network, for anomaly-resilient real-time temporal QoS prediction. To address data deficiency, HCTN combines non-negative matrix decomposition with hypergraph-based collaborative filtering for richer, higher-order features. It handles data credibility by using a robust loss and a greysheep-aware time-sensitive strategy. For effective representation learning, HCTN integrates domain-specific and deep hypergraph-transformer-based features. The model outperformed existing methods in accuracy, with negligible inference latency, making it apt for real-time systems. In the future, we will explore energy-efficient ML service selection for sustainable deployment.

\bibliographystyle{IEEEtran}
\bibliography{references}

\appendices
\counterwithin{figure}{section}
\counterwithin{table}{section}

\section{GreySheep Discrepancy Index (GDI)}
\noindent 
As outlined in Section III-C1 of the main paper, the primary goal of introducing GDI is to determine how a user or service may exhibit unique characteristics that deviate from the general norms of other users and services. 
The GDI for a user $u_i$ (denoted as $\mathbb{G}^{t}(u_i)$) or a service $s_j$ (denoted as $\mathbb{G}^{t}(s_j)$) at time-step $t$ is a scalar value calculated by comparing the QoS invocation profile of $u_i$ (i.e., $\mathcal{Q}^t(i, .)$) or $s_j$ (i.e., $\mathcal{Q}^t(., j)$) with the overall QoS mean for $u_i$ (denoted as $\mu^t (u_i) = mean(\mathcal{Q}^t(i, .))$) or $s_j$ (denoted as $\mu^t (s_j) = mean(\mathcal{Q}^t(., j))$). This comparison incorporates individual means for each service when $u_i$ interacts with a service, or for each user when $s_j$ is invoked by a user. The mathematical definitions of $\mathbb{G}^{t}(u_i)$ and $\mathbb{G}^{t}(s_j)$ are provided in
Eq.s \ref{eq:gmu} and \ref{eq:gms}, respectively.
\begin{equation}\label{eq:gmu} \scriptsize
    \mathbb{G}^{t}(u_i) = \frac{\sum \limits_{s_j \in \mathcal{S}^t_i}\left(|\mathcal{Q}^{t}(i, j) - \mu^t(u_i) - \bar{\mu}^t(s_j)| \times \hat{\mathcal{N}}^{t}(s_j)\right)}{|\mathcal{S}^{t}_i|}
\end{equation}
\begin{equation}\label{eq:gms} \scriptsize
\mathbb{G}^{t}(s_j) = \frac{\sum \limits_{s_j \in \mathcal{U}^t_j}\left(|\mathcal{Q}^{t}(i, j) - \mu^t(s_j) - \bar{\mu}^t(u_i)| \times \hat{\mathcal{N}}^{t}(u_i)\right)}{|\mathcal{U}^{t}_j|}
\end{equation}
where, $\mathcal{S}^t_i$ represents the set of services invoked by $u_i$ at time-step $t$, and $\mathcal{U}^t_j$ represents the set of users that invoked service $s_j$ at $t$. The mean-centered QoS values for $u_i$ and $s_j$, denoted by $ \bar{\mu}^t(s_j)$ and $\bar{\mu}^t(u_i)$ respectively, are defined as follows:
\begin{equation}\label{eq:mean_service} \scriptsize
    \bar{\mu}^t(s_j) = \frac{\sum \limits_{u_i \in \mathcal{U}^t_j} \left(\mathcal{Q}^{t}(i, j) - \max \limits_{u_i \in \mathcal{U}^t_j}\mathcal{Q}^{t}(i, j) - \min\limits_{u_i \in \mathcal{U}^t_j}\mathcal{Q}^{t}(i, j)\right)} {{|\mathcal{U}^t_j| - 2}}
\end{equation}
\begin{equation}\label{eq:mean_user} \scriptsize
    \bar{\mu}^t(u_i) = \frac{\sum \limits_{s_j \in \mathcal{S}^t_i} \left(\mathcal{Q}^{t}(i, j) - \max \limits_{s_j \in \mathcal{S}^t_i}\mathcal{Q}^{t}(i, j) - \min\limits_{s_j \in \mathcal{S}^t_i}\mathcal{Q}^{t}(i, j)\right)}{|\mathcal{S}^t_i| - 2}
\end{equation}
Additionally, $\hat{\mathcal{N}}^t(u_i)$ and $\hat{\mathcal{N}}^t(s_j)$ represent the normalized standard deviation for $u_i$ and $s_j$ at time-step $t$, as shown in Eq.s \ref{eq:deviation_ui} and \ref{eq:deviation_si}, respectively. 
\begin{equation}\label{eq:deviation_ui} \scriptsize
        {\hat{\mathcal{N}}^t}(u_i) = 1 -  \left(\frac{\sigma^t(u_i) - \min\limits_{u_k \in {\mathcal{U}}} \sigma^t(u_k)} {\max\limits_{u_k \in {\mathcal{U}}} \sigma^t(u_k) - \min\limits_{u_k \in {\mathcal{U}}} \sigma^t(u_k)}\right)
\end{equation}
\begin{equation}\label{eq:deviation_si} \scriptsize
        {\hat{\mathcal{N}}^t}(s_j) = 1 - \left(\frac{\sigma^t(s_j) - \min\limits_{s_k \in {\mathcal{S}}} \sigma^t(s_k)} {\max\limits_{s_k \in {\mathcal{S}}} \sigma^t(s_k) - \min\limits_{s_k \in {\mathcal{S}}} \sigma^t(s_k)}\right)
\end{equation}
where, $\sigma^t(u_i) = sd(\mathcal{Q}^t(i, .))$ and $\sigma^t(s_j) = sd(\mathcal{Q}^t(., j))$ represent the standard deviation of the QoS vectors corresponding to $u_i$ and $s_j$ at time-step $t$, respectively.

A high value of $\hat{\mathcal{N}}^t(u_i)$ or $\hat{\mathcal{N}}^t(s_j)$ generally indicates a consistent QoS invocation pattern for $u_i$ or $s_j$ at time-step $t$. Therefore, when calculating the GDI of a user $u_i$ in relation to its individual service invocation $s_j$, we use $\hat{\mathcal{N}}^t(s_j)$ as a weight. This weight reflects the consistency of $s_j$ and helps to determine whether an abnormal value is attributable to $u_i$ or $s_j$. Similarly, the GDI for $s_j$ is calculated.
A high GDI value for $u_i$ or $s_j$ indicates a significant abnormality. 
Based on these GDI values, we labeled the greysheep users and services in the main paper. 


\section{Datasets}
\noindent
In this section, we provide more information about the datasets utilized in our experiments. We leverage two large-scale QoS datasets from the publicly available WSDREAM-2 \cite{WSDREAM} repository, namely, Response Time (RT) and Throughput (TP). 
These datasets capture QoS interactions for 142 users across 4500 web services over 64 time-steps, with a 15-minute interval between each time-step. 
These datasets are commonly used for time-aware QoS prediction \cite{tpmcf,DeepTSQP,GFEN,TUIPCC,RNCF,RTF}. 
We now discuss the characteristics of the datasets.

\begin{table*}[!h]
    \centering
    \caption{Range-wise analysis of QoS datasets}
        \begin{tabular}{l|l|cccccccc}
            \hline
            \multirow{2}{*}{RT (seconds)} & Range & (0.001, 0.5) & (0.5, 1) & (1, 2) & (2, 3) & (3, 5) & (5, 10) & (10, 15) & (15, 19.999) \\                              
            & Percentage & 56.44\% & 15.55\% & 8.89\% & 3.04\% & 3.51\% & 4.02\% & 1.80\% & 6.75\% \\ \hline

            \multirow{2}{*}{TP (kbps)} & Range & (3.6534e-5, 25) & (25, 50) & (50, 100) & (100, 250) & (250, 500) & (500, 1000) & (1000, 2500) & (2500, 6726.8335) \\              & Percentage & 90.66\% & 3.81\% & 2.57\% & 1.97\% & 0.69\% & 0.26\% & 0.0378\%  & 0.0008\% \\ \hline
            \end{tabular}
            \label{tab:rt_tp_range-wise}
\end{table*}

\begin{figure}[!b]
    \centering
    \scriptsize
    \begin{tikzpicture}
        \begin{axis}[
            width=0.85\linewidth,
            height=0.5\linewidth,
            xlabel={Services},
            ylabel={QoS Value},
            symbolic x coords={1, 2, 3, 4, 5, 6, 7, 8, 9, 10, 11, 12},
            xtick=data,
            legend style={
                at={(0.605,0.97)},
                anchor=north,
                legend columns=2,
                /tikz/every even column/.append style={column sep=-0.02cm},
                draw=none
            },
            legend image post style={scale=0.25},
            ymin=0.0, ymax=20.0,
            ymajorgrids=true,
            xmajorgrids=true,
            grid style=dashed,
            enlarge x limits=false,
        ]   

        \addplot+[color=blue, mark=*, thick] coordinates {
            (1, 0.138) (2, 0.986) (3, 0.316)	(4, 0.421)	(5, 0.121) (6, 0.27)
            (7, 0.045) (8, 0.187) (9, 0.201) (10, 1.394) (11, 0.165) (12, 0.51)
            };

        \addplot+[color=red, mark=square*, thick] coordinates {
           (1, 0.421) (2, 0.136)	(3, 1.042)	(4, 0.165)	(5, 0.387) (6, 0.427)
            (7, 0.09) (8, 0.277) (9, 0.045) (10, 0.198) (11, 0.123) (12, 0.063)
            };

        \addplot+[color=green, mark=diamond*, thick] coordinates {
            (1, 0.789) (2, 0.342)	(3, 2.466)	(4, 0.733)	(5, 0.929) (6, 0.413)
            (7, 0.701) (8, 0.184) (9, 0.723) (10, 0.412) (11, 3.9) (12, 0.225)
            };

        \addplot+[color=gray, mark=o, thick] coordinates {
            (1, 20.0) (2, 3.644) (3, 20.0)	(4, 18.791)	(5, 20.0) (6, 8.533)
            (7, 7.492) (8, 0.343) (9, 1.39) (10, 17.698) (11, 11.822) (12, 6.965)
            };

        \legend{NGS1, NGS2, NGS3, GS1}
        \end{axis}
    \end{tikzpicture}
    \caption{Example of a greysheep user}
    \label{fig:gs_ngs_plot}
\end{figure}

\textit{(i) Dataset Challenges:}  
In real-life scenarios, a user invokes very few services, eventually leading to situations where a large percentage of QoS data is unknown at a particular time slice, which causes high sparsity in the QoS log matrices. The sparsity is further increased when new users or services, often called cold-start, are added to the QoS ecosystem. The unknown QoS values and cold-start turn out to be data deficiency issues. The datasets may further have the issue of data credibility, where the data itself is unreliable due to the presence of outliers and greysheep (GS). The actual cause of generating outliers, however, is unknown, but it is led by network loads and service status. Moreover, the greysheep is known for its pattern of QoS invocation, which differentiates it from other users or services. Here, we exemplify the concept of GS using a subset of the RT dataset \cite{WSDREAM}. Fig. \ref{fig:gs_ngs_plot} depicts an instance of a GS user compared to three other non-GS (NGS) users across twelve services based on their response time values. Notably, the QoS pattern of GS1 significantly diverges from NGS1, NGS2, and NGS3. The presence of such users in the QoS logs complicates the learning process and leads to low QoS prediction performance. Therefore, handling such issues is essential for high-prediction accuracy.     

\textit{(ii) Dataset Distribution:} 
As reported in Section IV of the main paper, the value ranges in the RT and TP datasets 
are $[0.001, 19.999]$ and $[3.6534{\times}10^{-5}, 6726.8335]$, respectively. 
Table \ref{tab:rt_tp_range-wise} provides further details on the distribution of both datasets. 
Notably, both datasets are positively skewed, with most QoS values concentrated on the left side of the distribution. 
This asymmetry complicates the interpretation of complex features, necessitating non-linear models to capture the underlying patterns, which is the focus of our paper.

\section{Configuration of Hyperparameters}
\noindent
Table \ref{tab:parameter} details the hyperparameters used for experimental analysis in the main manuscript. We aim to optimize model performance across various configurations by systematically adjusting these hyperparameters {through an exhaustive grid search over the specified parameter ranges, conducted on the validation set using a five-fold cross-validation strategy. The configuration achieving the lowest average error in terms of MAE and RMSE was selected as the final model setting. During hyperparameter tuning, each configuration was trained for up to 20K epochs using an AdamW optimizer \cite{adamw} with a stepwise learning rate schedule. The learning rate was initialized at $10^{-3}$ and progressively reduced to facilitate smoother convergence and prevent premature stagnation. Early stopping ($\text{patience} = 250$) was employed to avoid overfitting and ensure stable training. A detailed sensitivity analysis of these hyperparameters is provided in Appendix~\ref{sec:impact_hyperparameters}}.

\begin{table}[!h] 
    \scriptsize
    \centering
    \caption{Hyperparameter settings of HCTN}
    \begin{tabular}{c l l}
        \hline
       Hyperparameters & Description & Values \\ \hline 
       $\tau$ & Time Window &  \{2, 4, 6, 8, 10, 12, 14\} \\ 
       $l$ & HCN layers &   \{1, 2, 3, 4\} \\ 
       $h_n$ & No. of heads &   \{1, 2, 3, 4, 5\} \\
       $d_k$ & Head size &   \{32, 64, 128, 256, 512\} \\ \hline
    \end{tabular}
    \label{tab:parameter}
\end{table}

\section{Comparison Methods}
\noindent
The experiments in the main manuscript are compared with the following representative methods:

\noindent
\emph{(i)} WSPred \cite{WSPRED}: This paper adopts a Tensor Factorization (TF) approach under the average QoS value regularization constraint for time-aware personalized QoS prediction. 

\noindent
\emph{(ii)} CTF \cite{CTF}: This method proposed an outlier-resilient TF method incorporating the Cauchy loss function. 

\noindent
\emph{(iii)} NNCP \cite{NNCP}: This paper proposes non-negative TF via canonical polyadic decomposition, multiplicative updating (MU) algorithm are leveraged to optimized the latent factors.

\noindent
\emph{(iv)} BNLFT \cite{BNLFT}: Extending NNCP, this method introduced linear biases to non-negative TF, and incorporates an alternating direction method along with MU for faster convergence.

\noindent
\emph{(v)} TUIPCC \cite{TUIPCC}: This paper proposed a similarity-based approach that prioritizes recent QoS values using temporal decay.

\noindent
\emph{(vi)} TRCF \cite{TRCF}: Combining ratio-based similarity and Pearson correlation, this method employs QoS averaging and deviation migration for temporal QoS prediction.

\noindent
\emph{(vii)} TaTruSR \cite{TaTruSR}: This method integrates a similarity-based approach with an entropy-based reliability estimation to effectively identify and exclude unreliable users.

\noindent
\emph{(viii)} DeepTSQP \cite{DeepTSQP}: Utilizing binary invocation and similarity-based features, this paper used multi-layer GRU-based QoS sequence learning.

\noindent
\emph{(ix)} GFEN \cite{GFEN}: This method uses a GAN model, an enhanced GRU generator, and a fully connected discriminator to mitigate the features in the probabilistic matrix factorization (PMF).

\noindent
\emph{(x)} RNCF \cite{RNCF}:  This paper employs a multi-layer GRU framework that uses one-hot encoded features as input.

\noindent
\emph{(xi)} SCATSF \cite{SCATSF}: This approach utilizes similarity-based neighbor selection that capture high-order features using a multi-layer deep network, and GRU for temporal QoS forecasting.

\noindent
\emph{(xii)} PLMF \cite{PLMF}: This technique uses personalized LSTM and matrix factorization (MF) to learn user and service latent factors for online QoS prediction.

\noindent
\emph{(xiii)} GMCL \cite{GMCL}: Combining tripartite graph modeling and contrastive learning, this method effectively captures both local and global behaviors improves robust features representation.

\noindent
\emph{{(xiv)}} 
{STGCN \cite{stgcn}: Originally proposed for traffic forecasting. STGCN introduces a spatio-temporal graph convolutional framework that simultaneously models both spatial and temporal dependencies from time-series sensor data.
}

\noindent
\emph{{(xv)}} 
{STGFT \cite{stgft}: The method was initially introduced as a spatio-temporal graph fusion transformer for modeling water-quality time-series in river networks.
}

\noindent
\emph{(xvi)} TPMCF \cite{tpmcf}: 
This method proposed a real-time outlier-resilient approach leveraging a GCN and an enhanced transformer encoder to captures temporal fluctuations.

\vspace{.15cm}
\noindent
\textbf{Positioning of HCTN:} In this paper, we propose HCTN, an end-to-end anomaly-resilient framework for real-time QoS prediction that utilizes high-order multi-granularity features through a hypergraph convoluted transformer encoder. This approach addresses various challenges, such as data deficiency, data credibility, and data representation in temporal QoS prediction. 
HCTN effectively mitigates various underlying challenges for highly accurate, time-sensitive QoS prediction, surpassing the limitations of conventional methods.

\begin{itemize}[left=0pt]
\item  
WSPred \cite{WSPRED}, CTF \cite{CTF}, NNCP \cite{NNCP}, and BNLFT \cite{BNLFT} employ TF-based approach to implicitly address sparsity and cold-start issues, while exploiting the triadic relationships among users, services, and time. \cite{WSPRED,NNCP,BNLFT} are sensitive to outliers due to their reliance on outlier-sensitive objective functions. Further, some of these methods incorporate regularization terms with the loss function to partially mitigate outlier effects. On the other hand, \cite{CTF} uses the Cauchy loss function to address outliers. However, all of these methods struggle to capture higher-order features and handle greysheep users. In contrast, our method, HCTN, employs a hypergraph convolution transformer network which not only enhances collaborative filtering by capturing more nuanced higher-order neighborhoods, it also captures fine-to-coarse-grained complex temporal dependencies. Further, to mitigate the greysheep problem, it leverages local features derived from QoS distributions to enhance the performance of greysheep instances. Thereby, demonstrating superior prediction performance compared to these methods.

\item 
TUIPCC \cite{TUIPCC}, TRCF \cite{TRCF}, and TaTruSR \cite{TaTruSR} 
 primarily rely on similarity computation and weighted averaging. These approaches, despite incorporating domain knowledge, often struggle with data deficiency, data credibility, and limited utilization of higher-order features. Although \cite{TaTruSR} introduces an entropy-based mechanism to identify unreliable users, it partially addresses the greysheep issue by eliminating them during evaluation rather than mitigating the root causes. In contrast, HCTN not only identifies such anomalous users (and services) but also enhances their representation through dedicated feature learning, effectively improving robustness and expressiveness.

\item 
DeepTSQP \cite{DeepTSQP}, GFEN \cite{GFEN}, RNCF \cite{RNCF}, SCATSF \cite{SCATSF}, and PLMF \cite{PLMF} 
utilize GRU/ LSTM to capture complex features effectively. While \cite{DeepTSQP,RNCF,GFEN,PLMF} rely on outlier-sensitive loss functions, \cite{SCATSF} adopts the Huber loss to partially mitigate the impact of outliers. Further, \cite{DeepTSQP} leverages similarity and binary invocation features, \cite{RNCF} uses ID embeddings, \cite{GFEN} applies PMF, \cite{SCATSF} incorporates contextual user-service features, and \cite{PLMF} employs latent MF features. Although some of these methods address a few more challenges, such as domain knowledge integration, data sparsity, and cold start, they still struggle with greysheep behavior and modeling long-range dependencies. Conversely, our method addresses both limitations effectively. Specifically, a dedicated greysheep mitigation module is proposed to identify and address the issue faced by such instances. Furthermore, to model the long-range dependencies in HCTN, we adopt a fine-to-coarse transformer-based module, leading to superior performance over these methods.

\item 
GMCL \cite{GMCL} introduced a tripartite graph-based learning framework trained with contrastive and L1 loss, which effectively addresses the outlier issue and captures the higher-order features. However, this approach does not explicitly address the cold start and greysheep problem and further relies on ID embeddings, resulting in declined prediction performance compared to HCTN, which is specifically designed to handle these issues effectively.

\item
{In contrast to 
STGCN \cite{stgcn} and STGFT \cite{stgft}, which were originally developed for traffic forecasting and water-quality prediction, respectively, by modeling spatio–temporal dependencies over homogeneous graphs constructed from physical sensors or river monitoring stations, we adapt these frameworks to the time-aware QoS prediction setting by applying them separately on user and service graphs. The node embeddings obtained from message passing on both graphs are concatenated row-wise and fed into an MLP, followed by an inner product to estimate the QoS matrix.
STGCN \cite{stgcn} employs first-order graph convolutions and temporal gated CNNs, which restrict spatial modeling to pairwise edges and limit temporal modeling to fixed-size receptive fields; as a result, it cannot capture multi-way collaborative structures or long-range QoS dynamics. 
STGFT \cite{stgft} improves temporal attention and introduces adaptive adjacency matrices but still relies on conventional graph structures that assume stationary spatial relationships, an assumption mismatched to the latent, irregular, and highly heterogeneous interaction patterns in QoS tensors. 
In contrast, HCTN introduces a hypergraph-based spatial module capable of modeling higher-order user–service co-invocation relationships beyond pairwise connectivity, allowing it to uncover richer collaborative dependencies even when direct observations are sparse. Furthermore, by integrating greysheep detection and local pattern adaptation, HCTN explicitly accounts for atypical or noisy interaction behaviors that STGCN and STGFT treat as unstructured noise. Its transformer-based temporal granularity module further captures fine-to-coarse temporal variations that conventional temporal convolution or single-level attention cannot model effectively. Theoretically, the combination of (i) higher-order hypergraph message passing, (ii) anomaly-aware feature refinement, and (iii) multi-resolution temporal attention yields a more expressive function class capable of approximating non-linear triadic relationships among users, services, and time. This increased representational capacity allows HCTN to exploit the same input features more effectively than STGCN and STGFT, ultimately leading to its superior performance on WSDREAM.}


\item 
TPMCF \cite{tpmcf} combines a GCN-based framework and transformer encoder architectures to capture spatial and temporal patterns. However, separate training of these two distinct architectures increases the number of parameters and prolongs training time. Although it uses a robust loss function to handle outliers, it fails to address the greysheep issue. In contrast, HCTN enjoys an end-to-end deep framework that incorporates a hypergraph collaborative filtering, eliminating the need for separately trained components and providing a sparsity-resilient solution to handle anomalous instances effectively.
\end{itemize}

\section{{Model Complexity Analysis}}
\noindent
{This section presents a detailed complexity analysis of HCTN regarding model size and computational efficiency. We first quantify the implementation-level complexity by reporting the total number of trainable parameters and estimating the floating-point operations (FLOPs) required per forward pass. We then complement these practical measurements with an asymptotic analysis of each major module to characterize HCTN’s theoretical computational behavior. Together, these results provide a comprehensive assessment of HCTN’s resource requirements and demonstrate its suitability for deployment under diverse hardware and memory constraints.}


\subsection{{Memory Efficiency and Computational Complexity}}
\noindent
{
Table~\ref{tab:param_flops} summarizes the computational complexity of HCTN in comparison with two strong QoS prediction baselines, CTF~\cite{CTF} and TPMCF~\cite{tpmcf}, by reporting both the number of trainable parameters and the floating-point operations (FLOPs) required per forward pass. The results indicate that HCTN achieved an effective balance between representational capacity and computational efficiency. Specifically, HCTN contains 551.56K parameters, an 8.6$\times$ reduction relative to TPMCF, while requiring only 27.95 GFLOPs, which is 4.5$\times$ lower than TPMCF’s computational cost.
While CTF attained the smallest model size and the lowest FLOPs, its prediction accuracy on the WSDREAM QoS benchmark (Tables III-IV of the main paper) remained noticeably lower than that of HCTN. This suggests that CTF’s highly compact design, although computationally attractive, may limit its ability to capture the richer spatio–temporal and collaborative patterns required for accurate QoS estimation. In contrast, HCTN offers a more favorable accuracy–efficiency trade-off: it delivers substantially improved prediction performance with only a modest increase in computation. Overall, the results demonstrate that HCTN remains computationally lightweight while providing strong predictive capability, making it suitable for large-scale, time-aware QoS prediction scenarios where both accuracy and deployability are essential.}

%
\begin{table}[!t]
    \centering
    \caption{Model parameter counts and FLOPs}
    \begin{tabular}{l c c} \hline
        Method & Params (K) & FLOPs (G)  \\ \hline
        CTF \cite{CTF} & 0.7059 & 1.8403 \\
        TPMCF \cite{tpmcf} & 4765.3880 & 126.6281 \\ \hline
        HCTN & 551.5640 & 27.9539 \\ \hline
    \end{tabular}
    \label{tab:param_flops}
\end{table}
\subsection{{Asymptotic Complexity Analysis}}
\begin{table}[!b]
    \centering
    \caption{Symbols used in complexity analysis}
    \begin{tabular}{c|l} \hline
        Symbols & Representations \\ \hline
         \(n\) & Number of users \\
         \(m\) & Number of services \\
         \(\tau\) & Number of past time steps \\
         \(l\) & Number of hypergraph-convolution layers \\
         \(e_t\) & Number of QoS interactions at time \(t\) \\ 
         \(h\) & Number of attention heads in MHA \\
         \(d\) & Transformer input dimension \\ \hline
    \end{tabular}
    \label{tab:hctn_bigoh}
\end{table}
\noindent
{Let, \(n\) and \(m\) denote the number of users and services, respectively, and \(N=n{+}m\) the total number of nodes. 
Let \(\tau\) represents the number of temporal snapshots, and \(E=\sum_{t=1}^{\tau} e_t\) denote the total number of observed QoS interactions across all \(\tau\) time steps. 
The asymptotic computational cost of each component of HCTN is summarized below:
\begin{itemize}[left=0pt]
    \item 
    \emph{Non-negative Matrix Decomposition (NMD):}  
    Decomposing each \(Q^t\!\in\!\mathbb{R}^{n\times m}\) into latent matrices  
    \(X_u^t\!\in\!\mathbb{R}^{n\times f_1}\) and \(X_s^t\!\in\!\mathbb{R}^{m\times f_1}\) costs 
    \(\mathcal{O}(n m f_1)\) per iteration.  
    Over \(I_{\mathrm{NMD}}\) iterations and \(\tau\) time steps:
    \({\mathcal{O}(I_{\mathrm{NMD}} \cdot \tau \cdot n m f_1)}.
    \)
    \item 
    \emph{Hypergraph Collaborative Filtering Module (HCFM):}  
    Each snapshot employs an HCN with one dense projection 
    \(\mathcal{O}(N f_1 f_2)\) and \(l\) hypergraph-convolution layers, each costing  
    \(\mathcal{O}(N f_2^2 + e_t f_2)\).  
    Summed across all \(\tau\) snapshots:
    \({\mathcal{O}(\tau N f_1 f_2 + l\tau N f_2^2 + l f_2 E)}.
    \)
    \item 
    \emph{Greysheep Mitigation Module (GMM):}  
    Computing local statistical features scales as \(\mathcal{O}(E)\),  
    while dense fusion across \(\tau\) time steps costs \(\mathcal{O}(\tau N f_2^2)\):  
    \({\mathcal{O}(E + \tau N f_2^2)}.
    \)
    \item 
    \emph{Temporal Granularity Extraction Module (TGEM):}  
    The multi-head attention (E-block) requires \(\mathcal{O}(N\tau^2 f_2)\),  
    and the subsequent dense transformations (T- and F-blocks) add \(\mathcal{O}(N\tau f_2^2)\):  
    \({\mathcal{O}(N\tau^2 f_2 + N\tau f_2^2)}.
    \)
    \item 
    \emph{Collaborative QoS Prediction Module (CQPM):}  
    Normalization and pooling incur \(\mathcal{O}(N\tau f_2)\),  
    dense projections contribute \(\mathcal{O}(N f_2 f_3 + N f_3 f_4)\),  
    and final QoS reconstruction \(U S^\top\) adds \(\mathcal{O}(n m f_4)\):  
    \(
    {\mathcal{O}(N\tau f_2 + N f_2 f_3 + N f_3 f_4 + n m f_4)}.
    \)
\end{itemize}
}

\noindent 
{\emph{Overall complexity:}  
Combining all components, the total asymptotic cost of HCTN is:
    \[
        \begin{aligned}
            \mathcal{O}~\!\Big(&
            I_{\mathrm{NMD}} \cdot \tau \cdot n m f_1
            + l\tau N f_2^2 + l f_2 E \\
            &\qquad
            +\, N\tau^2 f_2 + N\tau f_2^2
            + n m f_4
            \Big).
        \end{aligned}
    \]
In sparse QoS settings (\(E\!\ll\!N^2\)) and small \(\tau\), the runtime is dominated by 
the hypergraph convolution term \(\mathcal{O}(l f_2 E)\) and the dense transformation term \(\mathcal{O}(N\tau f_2^2)\), 
ensuring linear scalability with the number of observed interactions and maintaining efficient training and inference.
}

\section{Impact of Cold-start}

\begin{figure}[!b] 
    \begin{subfigure}{0.21\textwidth} 
        \centering 
        \scriptsize
        \begin{tikzpicture}
            \begin{axis}[
                width=1.25\textwidth,
                height=0.85 \textwidth,
                ylabel={MAE},
                symbolic x coords={\shortstack{D1.2\\(CU)}, \shortstack{D1.2\\(CS)}, \shortstack{D1.2\\(CB)},\shortstack{D1.4\\(CU)}, \shortstack{D1.4\\(CS)},\shortstack{D1.4\\(CB)}},
                xticklabel style={rotate=75, anchor=east},
                xtick=data,
                ybar=1pt,
                legend style={at={(0.50,0.97)}, anchor=north,legend columns=5, /tikz/every even column/.append style={column sep=-0.05cm},
                draw=none},
                legend image post style={scale=0.4},
                ymin=0,ymax=1.5,
                bar width=1.7 pt,
                ymajorgrids=true,
                xmajorgrids=true,
                grid style=dashed
            ]

           \addplot [fill=blue, draw=none, draw opacity=0] coordinates {(\shortstack{D1.2\\(CU)},0.6765) (\shortstack{D1.2\\(CS)},0.6765) (\shortstack{D1.2\\(CB)},0.6765)(\shortstack{D1.4\\(CU)},0.5256) (\shortstack{D1.4\\(CS)},0.5256) (\shortstack{D1.4\\(CB)},0.5256)};
           
           \addplot [fill=red, draw=none, draw opacity=0] coordinates  {(\shortstack{D1.2\\(CU)},0.7510) (\shortstack{D1.2\\(CS)},0.8067 ) (\shortstack{D1.2\\(CB)},0.8050) (\shortstack{D1.4\\(CU)},0.5805) (\shortstack{D1.4\\(CS)},0.6158) (\shortstack{D1.4\\(CB)},0.6099)};
           
           \addplot [fill=gray, draw=none, draw opacity=0] coordinates {(\shortstack{D1.2\\(CU)}, 0.7525) (\shortstack{D1.2\\(CS)},0.8087) (\shortstack{D1.2\\(CB)},0.8541) (\shortstack{D1.4\\(CU)}, 0.5873) (\shortstack{D1.4\\(CS)},0.6196) (\shortstack{D1.4\\(CB)},0.6221)}; 
           
           \addplot [fill=green, draw=none, draw opacity=0] coordinates {(\shortstack{D1.2\\(CU)},0.7595) (\shortstack{D1.2\\(CS)},0.8600) (\shortstack{D1.2\\(CB)},0.9140) (\shortstack{D1.4\\(CU)},0.5923) (\shortstack{D1.4\\(CS)},0.6502) (\shortstack{D1.4\\(CB)},0.7119)}; 
           
           \addplot [fill=cyan, draw=none, draw opacity=0] coordinates {(\shortstack{D1.2\\(CU)},0.7889) (\shortstack{D1.2\\(CS)},0.8633) (\shortstack{D1.2\\(CB)},0.9397) (\shortstack{D1.4\\(CU)},0.5985) (\shortstack{D1.4\\(CS)},0.6509) (\shortstack{D1.4\\(CB)},0.7142)};    

            \legend{\scalebox{0.75}{$\xi$=0}, \scalebox{0.75}{$\xi$=5}, \scalebox{0.75}{$\xi$=10}, \scalebox{0.75}{$\xi$=15}, \scalebox{0.75}{$\xi$=20}}
            \end{axis}
        \end{tikzpicture}
        \caption{\scriptsize{D1 Dataset}}
         \label{fig:cs_rt}
    \end{subfigure}
    ~~
    \begin{subfigure}{0.21\textwidth} 
        \centering
        \scriptsize
        \begin{tikzpicture}
            \begin{axis}[
                width=1.25 \textwidth,
                height=0.85 \textwidth,
                ylabel={MAE},
                symbolic x coords={\shortstack{D2.2\\(CU)}, \shortstack{D2.2\\(CS)}, \shortstack{D2.2\\(CB)},\shortstack{D2.4\\(CU)}, \shortstack{D2.4\\(CS)},\shortstack{D2.4\\(CB)}},
                xticklabel style={rotate=75, anchor=east},
                xtick=data,
                ybar=1pt,
                legend style={at={(0.5,0.97)}, anchor=north,legend columns=5, /tikz/every even column/.append style={column sep=-0.05cm},
                draw=none},
                legend image post style={scale=0.40},
                ymin=2,ymax=8.0,
                bar width=1.7pt,
                ymajorgrids=true,
                xmajorgrids=true,
                grid style=dashed
            ]
 
           \addplot [fill=blue, draw=none, draw opacity=0] coordinates {(\shortstack{D2.2\\(CU)},3.5486) (\shortstack{D2.2\\(CS)},3.5486) (\shortstack{D2.2\\(CB)},3.5486)(\shortstack{D2.4\\(CU)},2.8671) (\shortstack{D2.4\\(CS)},2.8671) (\shortstack{D2.4\\(CB)},2.8671)};
           
           \addplot [fill=red, draw=none, draw opacity=0] coordinates  {(\shortstack{D2.2\\(CU)},3.6479) (\shortstack{D2.2\\(CS)},3.9635) (\shortstack{D2.2\\(CB)},3.9443) (\shortstack{D2.4\\(CU)},3.1539) (\shortstack{D2.4\\(CS)},3.5236) (\shortstack{D2.4\\(CB)},3.6499)};
           
           \addplot [fill=gray, draw=none, draw opacity=0] coordinates {(\shortstack{D2.2\\(CU)},3.8068) (\shortstack{D2.2\\(CS)},4.399) (\shortstack{D2.2\\(CB)},4.6851) (\shortstack{D2.4\\(CU)},3.2107) (\shortstack{D2.4\\(CS)},3.5473) (\shortstack{D2.4\\(CB)},4.2163)};
           
           \addplot [fill=green, draw=none, draw opacity=0] coordinates {(\shortstack{D2.2\\(CU)},3.8418) (\shortstack{D2.2\\(CS)},4.733) (\shortstack{D2.2\\(CB)},5.4062) (\shortstack{D2.4\\(CU)},3.5101) (\shortstack{D2.4\\(CS)},3.9467) (\shortstack{D2.4\\(CB)},4.4329)};
           
           \addplot [fill=cyan, draw=none, draw opacity=0] coordinates {(\shortstack{D2.2\\(CU)},4.023) (\shortstack{D2.2\\(CS)},5.0934) (\shortstack{D2.2\\(CB)},5.6886) (\shortstack{D2.4\\(CU)},3.5135) (\shortstack{D2.4\\(CS)},4.0506) (\shortstack{D2.4\\(CB)},4.8293)};

            \legend{\scalebox{0.75}{$\xi$=0}, \scalebox{0.75}{$\xi$=5}, \scalebox{0.75}{$\xi$=10}, \scalebox{0.75}{$\xi$=15}, \scalebox{0.75}{$\xi$=20}}
            \end{axis}
        \end{tikzpicture}
        \caption{\scriptsize{D2 Dataset}}
        \label{fig:cs_tp}
    \end{subfigure}
     \caption{Performance of HCTN on cold-start with varying $\xi$}
    \label{fig:coldstart_rt_tp}
\end{figure}

\noindent
{To assess the impact of cold-start conditions on HCTN, we simulate controlled cold-start scenarios by randomly selecting a fixed percentage $\xi \in \{5,10,15,20\}$ of users or services and removing all their historical QoS records. Three evaluation settings are constructed: (a) \emph{Cold-start Users (CU)}, where QoS histories for $\xi\%$ of users are removed; (b) \emph{Cold-start Services (CS)}, where QoS histories for $\xi\%$ of services are removed; and (c) \emph{Cold-start Both (CB)}, where QoS histories for $\xi\%$ of both users and services are removed. After applying the respective cold-start mask, the model is trained using the remaining valid entries following the same train–validation–test protocol described in Section~IV of the main paper. For each $\xi$ and each scenario, the complete training and evaluation pipeline is repeated five times with different random selections, and the mean performance across the five runs is reported.}
Fig. \ref{fig:coldstart_rt_tp} shows the results of these scenarios, and our key observations are summarized below:

\begin{table*}[!t]
    \centering
    \caption{Cold-start users and service comparison of HCTN with CTF and TPMCF}
    \begin{tabular}{c | c c c | c c c | c c c | c c c} \hline
        \multirow{2}{*}{$\xi$} & \multicolumn{3}{c|}{D1.2} & \multicolumn{3}{c|}{D1.4} & \multicolumn{3}{c|}{D2.2} & \multicolumn{3}{c}{D2.4} \\ \cline{2-13}
        & CTF & TPMCF & HCTN &  CTF & TPMCF & HCTN &  CTF & TPMCF & HCTN &  CTF & TPMCF & HCTN \\ \hline
        0 &  1.8908 & 0.8132 & 0.6765 & 1.5337 & 0.6159 & 0.5256 & 10.9274 & 5.7447 & 3.5486 & 8.6924 & 4.5136 & 2.8671\\
        5 & 1.9283 & 0.8490 & 0.8050 & 1.6072  & 0.6945 & 0.6099 & 11.7714 & 7.0479 & 3.9443 & 9.4462 & 5.4211 & 3.6499 \\
        10 & 1.9462 & 0.8993 & 0.8541 & 1.6173  & 0.7069 & 0.6221 &  12.7103 & 7.2463 & 4.6851 & 9.9722 & 5.5479 & 4.2163 \\
        15 & 1.9538 & 0.9419 & 0.9140 & 1.6235  & 0.7259 & 0.7119 & 12.6588 & 7.2771 & 5.4062 & 10.0847 & 5.6973 & 4.4329 \\
        20 & 1.9624 & 0.9590 & 0.9397 & 1.6303 & 0.7708 & 0.7142 & 12.8640 & 7.3544 & 5.6886 &  10.4121 & 5.9460 & 4.8293 \\ \hline
    \end{tabular}
    \label{tab:cs_sota}
\end{table*}

\begin{itemize}[left=0pt]
    \item[\emph{(i)}] As $\xi$ increased from 0 to 20 for all cases,  HCTN performance declined. Further, for the same value of $\xi$, the performance degradation was highest in the CB scenario, followed by CS; with CU showing the least decline possibly due to 
    the impact of service on the QoS value of a user-service pair is more prominent than that of a user. 
     
    \item[\emph{(ii)}] As the training density increased, performance increased for all scenarios \--- CU, CS, and CB, for the same value of $\xi$. This suggests that even with cold-starts in QoS logs, the effects can be reduced by increasing training density,  supplementing the matrix factorization to generate better latent features to handle cold-start. This justifies the requirements of matrix factorization features.  
    
\end{itemize}

\noindent
To show the effectiveness of HCTN in addressing the cold-start problem, we compare it with two prominent past methods, CTF \cite{CTF} and TPMCF \cite{tpmcf}, both handle temporal QoS prediction. 
CTF leverages tensor factorization to handle cold-start situations, while TPMCF uses non-negative matrix factorization to extract user/ service features, helping to alleviate the cold-start issue.

Table \ref{tab:cs_sota} presents a comparison of the CB scenario across various cold-start percentages, $\xi =\{0, 5, 10, 15, 20\}\%$, on all datasets. 
On average, the performance gains of the HCTN over TPMCF on D1 and D2 datasets are $8.01\%$ and $30.01\%$, respectively, which shows the effectiveness of HCTN over TPMCF. 
The performance gain of HCTN with respect to CTF is even higher, highlighting the benefits of utilizing higher-order features. This performance boost in HCTN is due to its effective feature utilization in cold-start situations and its ability to model non-linear features.

\section{Impact of Hyper-parameters} \label{sec:impact_hyperparameters}
\noindent
This section highlights the impact of different hyperparameters on the performance of HCTN.

\subsection{Impact of Time Window} 
\noindent
We explore the optimal number of time sequences needed for high QoS prediction performance. Fig.s \ref{fig:time_window_rt_tp}(a)-(b) show the experiments by varying the time window ($\tau$) from 2 to 12 with a step size of 2 for all datasets. The performance improves as $\tau$ increases for both datasets. At relatively lower $\tau$, performance deteriorates, which is due to not incorporating adequate temporal features essentially required for the high prediction performance. Further, incorporating more time slices enhances QoS prediction up to a certain $\tau$. Specifically, for the D1.2 and D1.4 datasets, we achieved the highest prediction performance for $\tau=10$ and $\tau=8$, respectively. Similarly, for D2 datasets, a common $\tau=10$ achieve the competitive performance. This experiment highlights that while a sufficient number of time slices is crucial for improving model performance by offering more time information, excessively large $\tau$ can negatively impact the model.

\begin{figure}[!t]
    \centering
    \begin{subfigure}{0.21\textwidth} 
        \centering
        \scriptsize
        \begin{tikzpicture}
            \begin{axis}[
                width=1.1\textwidth,
                height=0.80 \textwidth,
                xlabel={Time Window ($\tau$)},
                ylabel={MAE},
                symbolic x coords={2, 4, 6, 8, 10, 12},
                xtick=data,
                legend style={at={(0.5,0.97)}, anchor=north,legend columns=2, /tikz/every even column/.append style={column sep=0.01cm},
                draw=none},
                legend image post style={scale=0.25},
                ymin=0.4,ymax=1.2,
                bar width=1.5pt,
                ymajorgrids=true,
                xmajorgrids=true,
                grid style=dashed
            ]   
            \addplot coordinates {(2, 0.8091) (4, 0.7246) (6, 0.6917) (8, 0.6838) (10, 0.6765) (12, 0.6876)};
            \addplot coordinates {(2, 0.6001) (4, 0.5456) (6, 0.5335) (8, 0.5227) (10, 0.5300) (12, 0.5347)};

            \legend{\scalebox{0.60}{D1.2 ($\lambda=0$)}, \scalebox{0.60}{D1.4 ($\lambda=0$)}}
            
            \end{axis}
        \end{tikzpicture}
        \caption{\scriptsize{D1 Dataset}}
         \label{fig:time_window_rt}
    \end{subfigure}
    ~~
    \begin{subfigure}{0.21\textwidth} 
        \centering
        \scriptsize
        \begin{tikzpicture}
            \begin{axis}[
                width=1.1\textwidth,
                height=0.80 \textwidth,
                xlabel={Time Window ($\tau$)},
                ylabel={MAE},
                symbolic x coords={2, 4, 6, 8, 10, 12},
                xtick=data,
                legend style={at={(0.5,0.97)}, anchor=north,legend columns=2, /tikz/every even column/.append style={column sep=0.01cm},
                draw=none},
                legend image post style={scale=0.25},
                ymin=2.0,ymax=6,
                ymajorgrids=true,
                xmajorgrids=true,
                grid style=dashed
            ]

             \addplot coordinates {(2, 4.3555) (4, 4.0643) (6, 3.9662) (8, 3.5486) (10, 4.0607) (12, 4.0821)};
             \addplot coordinates {(2, 3.5327) (4, 3.2862) (6, 3.2451) (8, 2.8671) (10, 3.3528) (12, 3.4067)};

            \legend{\scalebox{0.60}{D2.2 ($\lambda=0$)},\scalebox{0.60}{D2.4 ($\lambda=0$)}}
            \end{axis}
        \end{tikzpicture}
        \caption{\scriptsize{D2 Dataset}}
        \label{fig:time_window_tp}
    \end{subfigure}
    \caption{Impact of number of time-windows ($\tau$)}
    \label{fig:time_window_rt_tp}
\end{figure}


%
\subsection{Impact of HCN Layers}
\noindent
This experiment shows the utilization of higher-order features obtained using hypergraph collaborative filtering. Fig.s \ref{fig:hcn_layer_rt_tp}(a)-(b) illustrate the impact of the number of HCN layers ($l$) by varying from 1 to 5. We achieved competitive performance for $l=2$ and $l=3$ for D1 and D2, respectively. However, for higher $l$, performance deteriorates since, with increased hypergraph convolution layer, the node features propagation may aggregated to give indistinguishable node embedding. 

\begin{figure}[!h]
    \centering
    \begin{subfigure}{0.21\textwidth}
        \centering
        \scriptsize
        \begin{tikzpicture}
            \begin{axis}[
                width=1.1\textwidth,
                height=0.80 \textwidth,
                xlabel={No. of HCN Layers ($l$)},
                ylabel={MAE},
                symbolic x coords={1, 2, 3, 4, 5},
                xtick=data,
                legend style={at={(0.50,0.97)}, anchor=north,legend columns=2, /tikz/every even column/.append style={column sep=0.01cm},
                draw=none},
                legend image post style={scale=0.25},
                ymin=0.4,ymax=1,
                bar width=1.5pt,
                ymajorgrids=true,
                xmajorgrids=true,
                grid style=dashed
            ]   
            \addplot coordinates{(1, 0.7036) (2, 0.6765)	(3, 0.6934)	(4, 0.7058)	(5, 0.7155)};
            \addplot coordinates {(1, 0.5543) (2, 0.5227) (3, 0.5331) (4, 0.5569) (5, 0.5771)};
            \legend{\scalebox{0.60}{D1.2 ($\lambda=0$)}, \scalebox{0.60}{D1.2 ($\lambda=10$)},\scalebox{0.60}{D1.4 ($\lambda=0$)}, \scalebox{0.60}{D1.4 ($\lambda=10$)}}
            \legend{\scalebox{0.60}{D1.2 ($\lambda=0$)},\scalebox{0.60}{D1.4 ($\lambda=0$)}}
            \end{axis}
        \end{tikzpicture}
        \caption{\scriptsize{D1 Dataset}}
         \label{fig:hcn_layer_rt}
    \end{subfigure}
    ~~
    \begin{subfigure}{0.21\textwidth}
        \centering
        \scriptsize
        \begin{tikzpicture}
            \begin{axis}[
                width=1.1\textwidth,
                height=0.80 \textwidth,
                xlabel={No. of HCN Layers ($l$)},
                ylabel={MAE},
                symbolic x coords={1, 2, 3, 4, 5},
                xtick=data,
                legend style={at={(0.50,0.97)}, anchor=north,legend columns=2, /tikz/every even column/.append style={column sep=0.01cm},
                draw=none},
                legend image post style={scale=0.25},
                ymin=2,ymax=5,
                ymajorgrids=true,
                xmajorgrids=true,
                grid style=dashed
            ]
            \addplot  coordinates {(1,  3.6684) (2, 3.5778) (3, 3.5486) (4, 3.6699) (5, 3.7063)};
            \addplot  coordinates {(1,  2.8715) (2, 2.8692) (3, 2.8671) (4, 2.8773) (5, 2.8965)};

            \legend{\scalebox{0.60}{D2.2 ($\lambda=0$)},\scalebox{0.60}{D2.4 ($\lambda=0$)}}
            \end{axis}
        \end{tikzpicture}
        \caption{\scriptsize{D2 Dataset}}
        \label{fig:hcn_layer_tp}
    \end{subfigure}
    \caption{Impact of no. of HCN layers ($l$)}
    \label{fig:hcn_layer_rt_tp}
\end{figure}


\subsection{Impact of Number of Heads and Head Size in MHA}
\noindent
To optimize the parameters in multi-head attention (MHA), we investigate the sufficient requirements of the number of heads ($h_n$) and head size ($d_k$) affecting model performance. Our key observations are as follows.

\begin{itemize}[left=0pt]
    \item[\emph{(i)}] 
Utilizing multiple heads in MHA enables the capture of multi-dimensional information from the features. This experiment intends to find the optimal number of heads that balance prediction performance with the number of learnable parameters, since an increment in the number of heads also increases the learnable parameters. 
Fig.s \ref{fig:head_head_size_rt_tp}(a)-(b) display the impact of $h_n$ for all datasets. 
We achieved the best performance for $h_n=3$ and $h_n=6$ for D1.2 and D1.4, respectively.   
However, $h_n=1$ for D2.2, and $h_n=3$ for D2.4 worked better.

\item[\emph{(ii)}] The head size refers to the dimensionality of each attention head in MHA. Specifically, it controls the size of features each attention head incorporates. For a fixed input features dimension, a smaller head size captures finer details, while larger heads capture broader information. To assess the head size, we vary $d_k$ in powers of $2$, ranging from $2^5$ to $2^9$. 
    Fig.s \ref{fig:head_head_size_rt_tp}(c)-(d) illustrate the performance with different $d_k$ for all datasets. We achieved optimal performance for $d_k = 2^7 = 128$ across all datasets. 
\end{itemize}

\begin{figure}[!t]
    \centering
    \begin{subfigure}{0.21\textwidth}
        \centering
        \scriptsize
        \begin{tikzpicture}
            \begin{axis}[
                width=1.1\textwidth,
                height=0.80 \textwidth,
                xlabel={No. of Heads ($h_n$)},
                ylabel={MAE},
                symbolic x coords={1, 2, 3, 4, 5, 6, 7, 8},
                xtick=data,
                legend style={at={(0.50,0.97)}, anchor=north,legend columns=2, /tikz/every even column/.append style={column sep=0.01cm},
                draw=none},
                legend image post style={scale=0.25},
                ymin=0.4,ymax=0.9,
                bar width=1.5pt,
                ymajorgrids=true,
                xmajorgrids=true,
                grid style=dashed
            ]
            \addplot coordinates {(1, 0.6928) (2, 0.6795) (3, 0.6765) (4, 0.6902) (5, 0.6900) (6, 0.6811) (7, 0.6836) (8, 0.7066)};
            \addplot coordinates {(1, 0.5274) (2, 0.5348) (3, 0.5227) (4, 0.5301) (5, 0.5337) (6, 0.5227) (7, 0.5304) (8, 0.5318)};
            
            \legend{\scalebox{0.60}{D1.2 ($\lambda=0$)}, \scalebox{0.60}{D1.4 ($\lambda=0$)}}

            \end{axis}
        \end{tikzpicture}
        \caption{\scriptsize{D1 Dataset}}
         \label{fig:heads_rt}
    \end{subfigure}
    ~~
   \begin{subfigure}{0.21\textwidth}
        \centering
        \scriptsize
        \begin{tikzpicture}
            \begin{axis}[
                width=1.1\textwidth,
                height=0.80 \textwidth,
                xlabel={No. of Heads ($h_n$)},
                ylabel={MAE},
                symbolic x coords={1, 2, 3, 4, 5, 6},
                xtick=data,
                legend style={at={(0.50,0.97)}, anchor=north,legend columns=2, /tikz/every even column/.append style={column sep=0.01cm},
                draw=none},
                legend image post style={scale=0.25},
                ymin=2, ymax=5,
                ymajorgrids=true,
                xmajorgrids=true,
                grid style=dashed
            ]
            \addplot coordinates {(1, 3.5486) (2, 3.6121) (3, 3.7266) (4, 3.6338) (5, 3.5645) (6, 3.6287)};
            \addplot coordinates {(1, 2.9394) (2, 2.9107) (3, 2.8671) (4, 2.9166) (5, 2.9710) (6, 2.9025)};
            
            \legend{\scalebox{0.60}{D2.2 ($\lambda=0$)}, \scalebox{0.60}{D2.4 ($\lambda=0$)}}
            
            \end{axis}
        \end{tikzpicture}
        \caption{\scriptsize{D2 Dataset}}
        \label{fig:heads_tp}
    \end{subfigure}
   \begin{subfigure}{0.21\textwidth}
        \centering
        \scriptsize
        \begin{tikzpicture}
            \begin{axis}[
                width=1.1\textwidth,
                height=0.80 \textwidth,
                xlabel={Head Size ($d_k$)},
                ylabel={MAE},
                symbolic x coords={32, 64, 128, 256, 512},
                xtick=data,
                legend style={at={(0.50,0.97)}, anchor=north,legend columns=2, /tikz/every even column/.append style={column sep=0.01cm},
                draw=none},
                legend image post style={scale=0.25},
                ymin=0.4, ymax=1,
                ymajorgrids=true,
                xmajorgrids=true,
                grid style=dashed
            ]
            
            \addplot  coordinates {(32, 0.7184) (64,  0.7107) (128, 0.6765) (256, 0.6987)  (512, 0.7069)};
            \addplot  coordinates {(32, 0.5724) (64, 0.5528) (128, 0.5227) (256, 0.5491) (512, 0.5639)}; 

            \legend{\scalebox{0.60}{D1.2 ($\lambda=0$)}, \scalebox{0.60}{D1.4 ($\lambda=0$)}}
            
            \end{axis}
        \end{tikzpicture}
        \caption{\scriptsize{D1 Dataset}}
        \label{fig:head_size_rt}
    \end{subfigure} 
    ~~
    \begin{subfigure}{0.21\textwidth} 
        \centering
        \scriptsize
        \begin{tikzpicture}
            \begin{axis}[
                width=1.1\textwidth,
                height=0.80 \textwidth,
                xlabel={Head Size ($d_k$)},
                ylabel={MAE},
                symbolic x coords={32, 64, 128, 256, 512},
                xtick=data,
                legend style={at={(0.50,0.97)}, anchor=north,legend columns=2, /tikz/every even column/.append style={column sep=0.01cm},
                draw=none},
                legend image post style={scale=0.25},
                ymin=2, ymax=5,
                ymajorgrids=true,
                xmajorgrids=true,
                grid style=dashed
            ]
            \addplot coordinates {(32, 3.6388) (64, 3.5658) (128, 3.5486) (256, 3.5491) (512, 3.5945)};
            \addplot coordinates {(32, 2.9006) (64, 2.8713) (128, 2.8671) (256, 2.9060) (512, 2.9205)};

            \legend{\scalebox{0.60}{D2.2 ($\lambda=0$)}, \scalebox{0.60}{D2.4 ($\lambda=0$)}}
            
            \end{axis}
        \end{tikzpicture}
        \caption{\scriptsize{D2 Dataset}}
        \label{fig:heads_size_tp}
    \end{subfigure}
\caption{ (a)-(b) Impact of no. of heads, and (c)-(d) head size}
    \label{fig:head_head_size_rt_tp}
\end{figure}

\section{Model Deployability}
\noindent
{We analyze the deployability of HCTN by measuring its inference latency across a diverse set of hardware configurations, ranging from mid-tier consumer CPUs to high-end workstation processors. 
As shown in Table~\ref{tab:inference_time_different_machine}, HCTN consistently achieved low inference times, with mean latency ranging from $2.12{\times}10^{-6}$~second to $2.58{\times}10^{-5}$~second across all machines. 
These results demonstrate that HCTN introduces negligible computational overhead and delivers real-time predictions without requiring GPU acceleration. 
The low variance in latency further indicates stable execution behavior across different memory capacities and core counts. Overall, the study confirms that HCTN is an efficient, lightweight, and easily deployable in practical QoS-aware service platforms under resource-constrained environments.}

\begin{table}[!h]
    \centering
    \caption{HCTN Inference time on different machines}
    \begin{adjustbox}{width=0.45\textwidth}
        \begin{tabular}{l|c|c|c} \hline
            \multirow{2}{*}{Processor Configuration} & \multirow{2}{*}{Core} & \multirow{2}{*}{RAM} & Inference Time (second) \\ 
            &&& (mean  $\pm$  stdev) \\ \hline

            10$^{\text{th}}$ Gen Intel® Core™ i7-10700@4.80 GHz & 16 & 7.53 GB & 2.58e-5 $\pm$ 1.19e-6  \\

            13$^{\text{th}}$ Gen Intel® Core™ i5-1335U@4.60 GHz & 12 & 15.30 GB & 1.67e-5 $\pm$ 1.66e-6 \\
            
            AMD Ryzen-9 7950X@5.70 GHz & 16 & 32.00 GB & 2.12e-6 $\pm$ 6.24e-8\\
             
            13$^{\text{th}}$ Gen Intel® Core™ i9-13900H@5.40 GHz & 20	& 30.98 GB & 2.01e-5 $\pm$ 2.30e-6 \\
             
            12$^{\text{th}}$ Gen Intel® Core™ i7-12700@4.90 GHz & 20	& 62.49 GB & 1.06e-5 $\pm$ 1.64e-7 \\
            
            12$^{\text{th}}$ Gen Intel® Core™ i7-12700@4.90 GHz & 20	& 125.49 GB & 1.10e-5 $\pm$ 1.35e-7 \\

            2$^{\text{nd}}$ Intel® Xeon Gold 6226R CPU@3.90GHz & 32 & 188.35 GB & 2.43e-5 $\pm$ 1.05e-7 \\ \hline
        \end{tabular}
        
    \end{adjustbox}
    \label{tab:inference_time_different_machine}
\end{table}


\section{Supporting Results for the Main Manuscript}

\noindent
In this subsection, we have moved some less prominent baseline methods (refer to Table \ref{tab:soa_table}) for the sake of completeness, while incorporating the most recent and highly relevant SOTA methods into the main manuscript.

\begin{table}[!ht]
    \centering
     \scriptsize 
        \caption{Comparison of HCTN with past methods}
        \adjustbox{max width=0.85\linewidth}{%
        \begin{tabular}{l | c | c c c c c | c c c c c}
        \hline 
        \multirow{2}{*}{Methods} & Loss & \multicolumn{5}{c|}{MAE} & \multicolumn{5}{c}{RMSE} \\ \cline{3-7} \cline{8-12}
        & Function & D1.1 & D1.2 & D1.3 & D1.4 & D1.5 & D1.1 & D1.2 & D1.3 & D1.4 & D1.5 \\ \hline 
        TF & MSE & 2.9184  & 2.7888 & 2.500 & 2.4800 & 2.2127 & 4.7508 & 4.5696 & 4.4100 & 4.3900 & 4.0169  \\  
        
        AMF \cite{AMF} & MRE & 1.6328 & 1.5766 & 1.5527	& 1.5391 & - & - & - & - & - & -  \\ 
 
        \textbf{HCTN} & \textbf{Cauchy} & \textbf{0.8453} & \textbf{0.6765} & \textbf{0.6133} & \textbf{0.5256} & \textbf{0.2237} & \textbf{2.4486} & \textbf{2.2336} & \textbf{2.1074} & \textbf{1.9754} & \textbf{1.2853} \\ \hline
    
        &  & D2.1 & D2.2 & D2.3 & D2.4 & D2.5 & D2.1 & D2.2 & D2.3 & D2.4 & D2.5 \\ \hline
        TF & MSE & 8.7997  & 8.5080 & 8.400 & 8.350 & 7.8045 & 39.5133 & 39.2792 & 39.3000 & 39.2800 & 38.6964  \\  
        AMF \cite{AMF} & MRE & 5.0942 & 4.6984 & 4.5398	& 4.4610 & - & 37.0818 & 36.0874 & 35.6750 & 35.5216 & -   \\ 
         \textbf{HCTN} & \textbf{Cauchy} & \textbf{3.9693} & \textbf{3.5486} & \textbf{3.3071} & \textbf{2.8671} & \textbf{0.7404} & \textbf{23.8219} & \textbf{21.3090} & \textbf{20.8300} & \textbf{18.8000} & \textbf{4.8526} \\ \hline
        \end{tabular}}
        \label{tab:soa_table}
\end{table}



\end{document}